\documentclass[11pt,letterpaper]{article}
\newcommand{\HiTArxivVersion}{}
\PassOptionsToPackage{table,dvipsnames}{xcolor}
\usepackage{hit_arxiv}

\usepackage{tabularx}
\usepackage{multirow}
\usepackage[table]{xcolor}

\makeatletter
\newcommand{\HiTrowshade}{%
  \smash{\rlap{%
    \color{gray!12}%
    \vrule width\columnwidth
           height\ht\@arstrutbox
           depth\dp\@arstrutbox
  }}%
}
\makeatother

\usepackage[normalem]{ulem}

\title{Hierarchical Diffusion Transformer for End-to-End Image Generation}
\author{%
    Xiaoyu Wu\textsuperscript{1,*,$\dagger$}\quad
    Yifei Wang\textsuperscript{2,*}\quad
    Weihang Guo\textsuperscript{2,*}\quad
    Chen Wei\textsuperscript{2}\\[2pt]
    Lydia E Kavraki\textsuperscript{2}\quad
    Niloofar Mireshghallah\textsuperscript{1}\quad
    Zhiwei Steven Wu\textsuperscript{1}%
}
\newcommand{\HiTAffiliations}{%
    \textsuperscript{1}Carnegie Mellon University\quad
    \textsuperscript{2}Rice University\\[3pt]
    \textsuperscript{*}Equal contribution.\quad
    \textsuperscript{$\dagger$}Project lead.%
}
\newcommand{\HiTContact}{%
    \HiTEmail{xiaoyuwu@andrew.cmu.edu},
    \HiTEmail{yw251@rice.edu},
    \HiTEmail{wg25@rice.edu},
    \HiTEmail{chen.wei@rice.edu}\\[1pt]
    \HiTEmail{kavraki@rice.edu},
    \HiTEmail{nmireshg@andrew.cmu.edu},
    \HiTEmail{zstevenwu@cmu.edu}%
}
\newcommand{\HiTResources}{%
    \begin{tabular}{@{}l@{\hspace{0.6em}}l@{}}
        \textbf{Project page:} & \url{https://nicholas0228.github.io/HiT-webpage/}\\
        \textbf{Code:} & \url{https://github.com/Nicholas0228/HiT-Released}\\
        \textbf{Models:} & \url{https://huggingface.co/Nicholas0228/HiT}
    \end{tabular}%
}
\hypersetup{
    pdftitle={Hierarchical Diffusion Transformer for End-to-End Image Generation},
    pdfauthor={Xiaoyu Wu, Yifei Wang, Weihang Guo, Chen Wei, Lydia E Kavraki, Niloofar Mireshghallah, Zhiwei Steven Wu},
    pdfsubject={HiT preprint}
}

\newcommand{\niloofar}[1]{}

\begin{document}

\begin{abstract}
Pixel-space diffusion learns visual representations and image
generation jointly in a single end-to-end training stage,
without a separately pretrained visual encoder.
However, plain pixel-space Transformers still trail leading
two-stage latent diffusion models in generation quality and
training efficiency.
We hypothesize that part of this gap arises from retaining
one token per patch throughout the backbone, which does not
explicitly encourage a visual hierarchy that summarizes
related content at progressively coarser levels.
To address this, we introduce Hierarchical Diffusion
Transformers (HiT), which progressively group related tokens
and represent each group with a single token.
The main Transformer bottleneck processes these compact
group tokens, and ungrouping restores features to their
original image positions.
A compression budget regulates the average compression ratio
across the batch, allowing group memberships and counts to
vary across images and diffusion timesteps during both
training and generation.
The model exhibits emergent semantic structure, learning
to group coherent objects and object parts without
segmentation labels.
In text-to-image generation, HiT further learns to group
caption words with image patches depicting the same concepts,
without region annotations.
HiT achieves \textbf{1.48 FID} on ImageNet $256\times256$
and reaches comparable generation quality to JiT using
approximately 40\% of its cumulative training FLOPs.
For text-to-image generation, HiT exceeds UniLiP's average
score across five benchmarks with roughly $6\times$ fewer
denoiser parameters and $3\times$ fewer forward FLOPs.
These results suggest that learning a visual hierarchy within
the generator breaks down the barrier to stronger image
generation without external visual pretraining.
\end{abstract}
\section{Introduction}
\label{sec:intro}

Latent diffusion models exploit a hierarchy between pixels and
compact visual representations~\citep{rombach2022ldm,peebles2023dit,
zheng2025rae,yu2025repa}.
A separately trained encoder organizes image information into
a feature space, where a diffusion model subsequently learns
its distribution. This hierarchy reduces the burden of direct
pixel modeling and supplies an organizational prior.
Despite their benefits for generation quality and convergence,
latent diffusion models inherit visual organization shaped
by encoder architectures and pretraining objectives outside
generative training.
Their hierarchical modeling therefore depends on a separate
visual representation learning stage.

Pixel-space diffusion removes this separate stage by learning
to generate images directly through end-to-end
training~\citep{hoogeboom2023simplediffusion,hoogeboom2025sid2,
chen2025pixelflow,yu2026pixeldit,li2025jit}.
This simplifies the training pipeline, yet matching leading
latent-space models in generation quality and convergence
speed remains challenging.
We hypothesize that part of this gap concerns how visual
information is organized.
During encoder pretraining, latent-space models learn a
visual hierarchy that progressively aggregates fine-grained
image information into compact, higher-level representations.
Pixel-space generators must instead develop such organization
within generative training itself.
This raises the central question of this paper:
\emph{Can a pixel-space model learn a useful visual hierarchy
within generative training, while retaining an end-to-end
visual pipeline?}

\begin{figure*}[t]
    \centering
    \includegraphics[width=\textwidth]{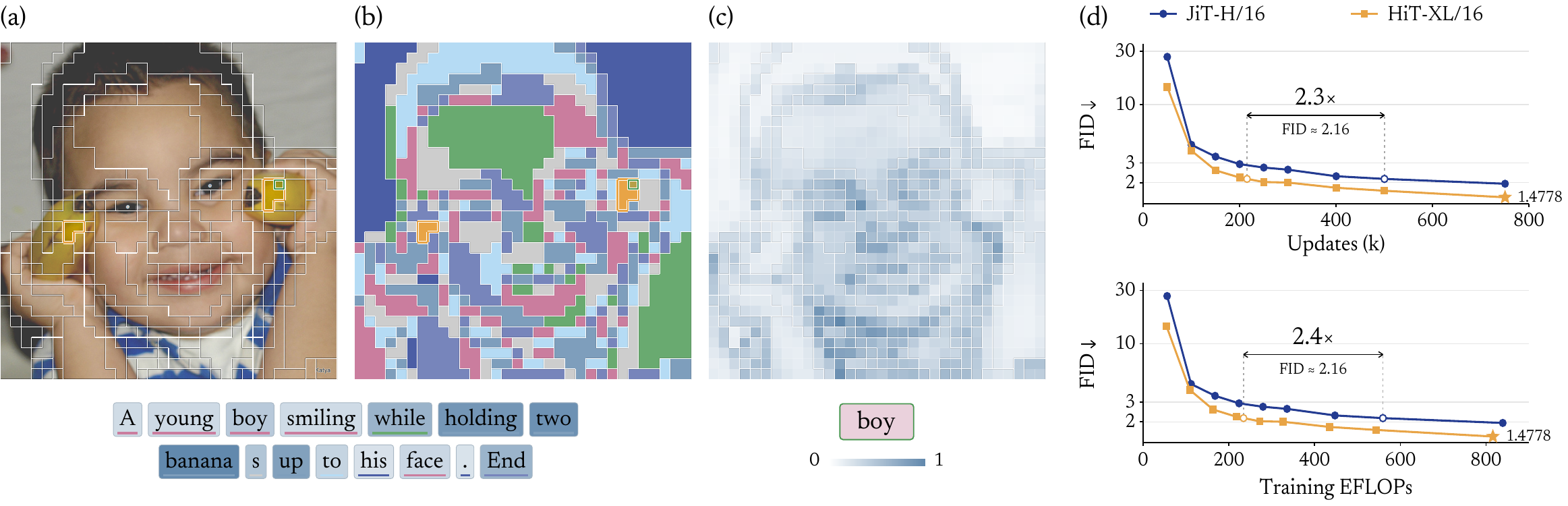}
    \caption{
        \textbf{HiT learns image--text associations without
        region annotations and converges faster.}
        (a) On a clean input ($t=1$), HiT groups patches
        with similar representations.
        (b) The corresponding group map.
        Spatially disconnected patches from both bananas
        share the yellow group.
        Darker blue caption words indicate stronger soft
        routing connections from the banana patch outlined
        in green.
        (c) For the selected word ``boy'', the image heatmap
        shows its group's soft pooling weights, emphasizing
        the boy's face and upper body.
        (d) FID-50k on a log scale for HiT-XL/16 and JiT-H/16
        on ImageNet-256, using guidance scales $2.6$ and $2.4$,
        respectively, plotted against optimizer updates (top)
        and estimated cumulative training compute (bottom).
        At FID $\approx2.16$, HiT uses $2.4\times$ less
        training compute and $2.3\times$ fewer updates.
    }
    \label{fig:hit-teaser}
\end{figure*}

Learning such a hierarchy requires deciding which information
can share a representation and where finer distinctions
are needed.
By contrast, a regular patch grid fixes this allocation
spatially, assigning the same number of tokens to equally
sized regions despite differences in their perceptually
relevant content~\citep{dieleman2025latents}.
This suggests learning the allocation itself rather than
prescribing it through a fixed spatial partition.
This motivates a learnable computational bottleneck that
jointly determines how image information is organized
and how many units are used to represent it.

To this end, we introduce Hierarchical Diffusion Transformers
(HiT), a new architecture that learns a visual hierarchy
within a single end-to-end pixel-space generator inspired by
H-Net~\citep{hwang2025hnet,cartesia2026hnet}.
Our central idea is to group image patch features at successive
stages and process the resulting group tokens in deeper
Transformer bottlenecks.
To realize this design, we use a learned router to group
these features and soft aggregation to form compact tokens,
so deeper Transformer stages operate on progressively
shorter sequences.
We then use ungrouping to return processed group features
to finer positions and refine them with the remaining
Transformer blocks.
We use an aggregate compression budget to regulate the average
token reduction across a batch, allowing the model to allocate
different numbers of groups according to each image's content
and diffusion timestep.

This architecture leads to several striking emergent behaviors.
Without segmentation data, the model learns to consolidate
large background regions into a few units and to group
coherent objects and object parts.
It also groups semantically related but spatially disconnected
regions.
For the same image, the model learns to reorganize group
memberships and adjust group counts across diffusion timesteps
as the noisy input evolves.

We further extend HiT to text-to-image
generation~\citep{wang2026minit2i} and observe additional
emergent behavior.
We let image and text features enter a shared grouping
hierarchy, allowing compact representations to combine
information across modalities.
As shown in Fig.~\ref{fig:hit-teaser}, HiT learns semantic
associations in both directions.
A selected banana patch assigns high soft routing weight
to the caption word ``banana''.
Conversely, the group containing ``boy'' pools features
from the boy's face and upper body.
These associations emerge even without explicit word--region
annotations.

Alongside these emergent behaviors, HiT achieves substantial
gains in generation quality and training efficiency.
On ImageNet $256\times256$, HiT reaches generation quality
comparable to JiT~\citep{li2025jit} with approximately half
as many optimizer updates and 40\% of the cumulative training
FLOPs.
After 600 epochs, HiT achieves \textbf{1.48 FID}, even surpassing pixel-space models that use external
visual supervision.
On text-to-image benchmarks, HiT substantially outperforms
the similarly sized MiniT2I-B/16 baseline~\citep{wang2026minit2i}
and surpasses UniLiP-3B's average score~\citep{tang2025unilip}
with roughly $6\times$ fewer denoiser parameters and
$3\times$ fewer forward FLOPs.
These results show that the barrier to stronger image generation is never the lack of external alignment, but the lack of a correct architecture that learns a visual hierarchy by itself.

\section{Related Work}
\label{sec:related}

\paragraph{Latent- and pixel-space image generation.}
Latent diffusion learns with compact representations supplied by pretrained autoencoders~\citep{rombach2022ldm,peebles2023dit}. Subsequent work further incorporates pretrained semantic encoders either as supervision for intermediate denoiser representations or as the generation space itself~\citep{yu2025repa,zheng2025rae}. These approaches benefit from visual organization learned outside the generative objective. Pixel-space methods instead pursue direct generation in one stage~\citep{hoogeboom2023simplediffusion,hoogeboom2025sid2,chen2025pixelflow,yu2026pixeldit}. JiT improves pixel-space generation through clean-image
prediction, while MiniT2I combines this formulation with
joint image--text attention for text-conditioned
generation~\citep{li2025jit,wang2026minit2i}. In these models, the architecture fixes the number and
spatial layout of image tokens at each stage.

Unlike these pixel-space generators with fixed token grids,
HiT learns which image features to group and how many group
tokens each input needs.
This learned hierarchy improves generation quality and
training efficiency within a single end-to-end training stage.

\paragraph{Token grouping and compression.}
Hierarchical sequence models concentrate computation on compact
representations.
BLT and H-Net form adaptive chunks through entropy-based
segmentation and learned boundary prediction,
respectively~\citep{pagnoni2025blt,hwang2025hnet}.
Both form chunks along the input sequence and process fewer
units at deeper levels.
Cartesia's subsequent vision experiments demonstrate the
potential of learned chunking for pixel-space image
generation~\citep{cartesia2026hnet}.

In vision, token reduction groups features directly across
image positions.
ToMe and ToMeSD merge similar tokens to accelerate visual
Transformers and pretrained diffusion
models~\citep{bolya2023tome,bolya2023tomesd}, while GroupViT
learns hierarchical semantic grouping through image--text
supervision~\citep{xu2022groupvit}.
These methods establish grouping as a useful mechanism for
computational efficiency and visual representation learning,
but use predefined token counts or reduction rates.

Unlike prior visual grouping methods with fixed token counts,
HiT learns grouping jointly with pixel-space denoising,
allowing both group memberships and counts to vary across
images and diffusion timesteps.
Unlike the original H-Net, which forms chunks along a causal
sequence, HiT uses affinity-based grouping over bidirectional
image features.
This yields a more flexible bottleneck in which spatially
disconnected regions can share a computational unit.

\section{Hierarchical Diffusion Transformer}
\label{sec:method}

HiT learns which image features can share a
representation and how many groups to retain for each input.
Through \emph{grouping}, we progressively form compact tokens
for deeper Transformer processing.
We then use \emph{ungrouping} to restore features to finer positions,
while skip connections preserve information from before
each grouping stage.

\subsection{Network Architectures}
\label{sec:architecture}

\begin{figure}[t]
    \centering
    \includegraphics[width=\columnwidth]{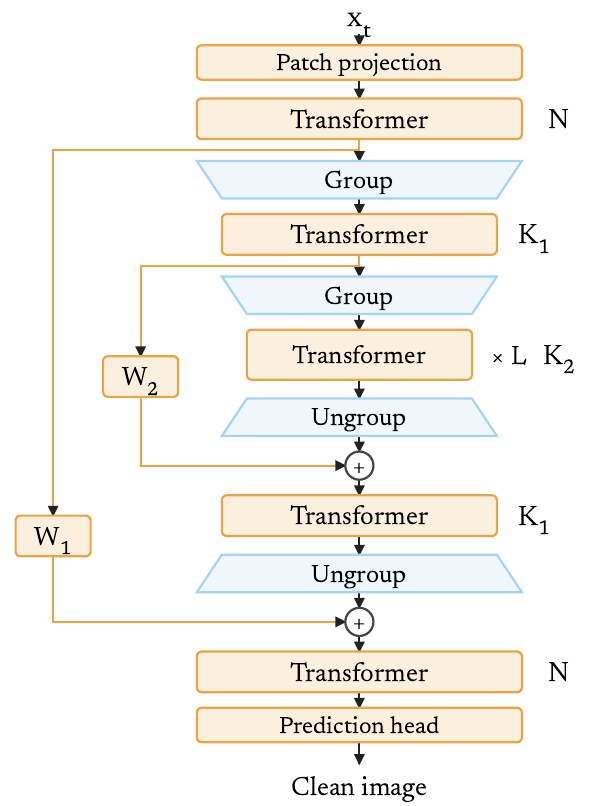}
\caption{
\textbf{Recursive architecture of HiT.}
Two grouping stages reduce the image-token count from
$N$ to $K_1$ to $K_2$, and ungrouping restores these counts
in reverse order.
The central $L$ Transformer blocks process the $K_2$
group tokens.
At each level, we add encoder features projected by
$W_\ell$ to the corresponding ungrouping output before
the next Transformer stage.
Narrower blocks indicate fewer tokens; the feature
dimension $d$ remains fixed.
}
    \label{fig:hit-architecture}
\end{figure}

\paragraph{Hierarchical backbone.}
We arrange Transformer blocks into nested encoder, compact network,
and decoder stages~\citep{hwang2025hnet}
(Fig.~\ref{fig:hit-architecture}).
A linear patch projection maps the noisy image to $N$ tokens.
Each encoder supplies features for grouping.
After compact processing, ungrouping restores the preceding token positions.
We add the projected encoder features before the corresponding decoder.
With $J$ grouping levels, the token counts follow
\begin{equation}
N\rightarrow K_1\rightarrow\cdots\rightarrow K_J
\rightarrow\cdots\rightarrow K_1\rightarrow N.
\end{equation}
Here, $K_\ell$ is the group count at level $\ell$.
Each level has its own router and skip projection $W_\ell$.
Transformer blocks preserve the token count and feature width.
At the deepest level, $L$ consecutive blocks process the compact
sequence without further grouping.
A final linear head maps image features to the predicted clean image,
following JiT~\citep{li2025jit}.

\subsection{Grouping}
\label{sec:group-construction}
\label{sec:soft-routing}

\begin{figure}[t]
    \centering
    \includegraphics[width=\columnwidth]{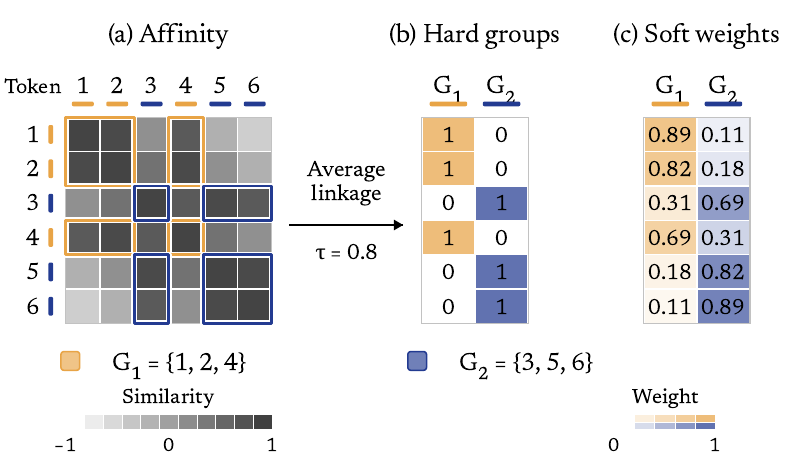}
    \caption{
    \textbf{Grouping in HiT.}
    All panels retain token order from 1 to 6.
    (a) Gray level shows the affinity $S_{ij}$.
    Colored boundaries mark affinities within each hard group.
    (b) Average linkage produces one hard membership per token.
    (c) Mean affinity to each hard group gives soft weights through
    a softmax. Yellow and blue identify the same groups across panels.
    Values are illustrative.
    }
    \label{fig:HiT}
\end{figure}

\paragraph{Affinity and hard grouping.}

At grouping stage $\ell$, let $H\in\mathbb{R}^{N\times d}$
denote the contextualized token features produced by the
preceding Transformer blocks, where $N$ is the current
number of tokens and $d$ is their feature width.
Each row $h_i$ represents one token entering the grouping
operation.
We learn two projections~\citep{hwang2025hnet}
and normalize each projected feature to unit length.
For token $i$, these give $q_i$ and $k_i$.
We compute the symmetric affinity
\begin{equation}
S_{ij}=\tfrac{1}{2}\left(q_i^\top k_j+q_j^\top k_i\right).
\label{eq:HiT-affinity}
\end{equation}
Starting from individual tokens, we merge the two groups with the
highest mean affinity across their members.
This average linkage procedure defines a hierarchy.
We retain merges with affinity at least $\tau$, producing
$K$ groups $\{G_1,\ldots,G_K\}$ (Fig.~\ref{fig:HiT}b).
Tokens can share a group even when their image positions are far apart.
With a fixed baseline threshold, the memberships and group count
change with the features at each denoising step.

\paragraph{Soft grouping and skip connections.}
The grouping procedure above is discrete, so gradients cannot
propagate through group assignments to the affinity projections.
To address this, we use soft weights to aggregate features
during grouping and reuse the same weights to restore token
features during ungrouping.

For each token, we compute its mean affinity to the members
of each hard group.
Applying a softmax over groups with temperature
$T_{\mathrm{r}}$ gives the weights
$A\in\mathbb{R}^{N\times K}$, where each row describes a
token's associations with the groups and sums to one
(Fig.~\ref{fig:HiT}c).
To construct a group feature, we normalize the corresponding
column of $A$ over all input tokens.
The resulting pooling matrix $P$ therefore forms each
compact token as a weighted average of the input features.

After the compact network $\mathcal{M}$ processes these group
tokens, we reuse $A$ to mix the group outputs back into the
original token positions.
A skip connection also carries features from before grouping,
preserving information around the compact network:
\begin{equation}
\begin{aligned}
Z &= P^\top H,\\
\widetilde{H} &= A\mathcal{M}(Z)+H W_\ell^\top.
\end{aligned}
\label{eq:HiT-stage}
\end{equation}
Here, $W_\ell\in\mathbb{R}^{d\times d}$ is a learned skip
projection initialized to the identity.
During backpropagation, we treat the hard partition as
constant, while gradients reach the affinity projections
through the soft pooling and ungrouping weights.
Appendix~\ref{sec:grouping-details} provides the exact
weight definitions.

\subsection{Compression Loss}
\label{sec:training}

We regulate compression across the batch rather than fixing
the number of groups for each input.
This allows group counts to vary with image content and
diffusion timestep.

At grouping stage $\ell$, let $r>1$ denote the target
compression ratio and $\rho=1/r$ the target retained fraction.
We compute the actual retained fraction $f_{\mathrm h}$
by dividing the total group count by the total incoming
token count, aggregating both counts across the batch
and all devices.
Its differentiable estimate $f_{\mathrm s}$ uses soft
group-count estimates with the same denominator.
Each stage uses its actual incoming token population.
We apply the ratio loss~\citep{hwang2025hnet}:
\begin{equation}
\mathcal{L}_{\mathrm{comp}}^{(\ell)}
=1+\frac{(f_{\mathrm h}-\rho)(f_{\mathrm s}-\rho)}
             {\rho(1-\rho)}.
\label{eq:HiT-budget}
\end{equation}
We treat $f_{\mathrm h}$ as a constant during backpropagation,
so the loss updates routing features through $f_{\mathrm s}$.
When $f_{\mathrm h}>\rho$, the loss encourages more merging;
when $f_{\mathrm h}<\rho$, it encourages retaining more groups.

We sum the compression losses across grouping stages,
$\mathcal{L}_{\mathrm{comp}}
=\sum_\ell\mathcal{L}_{\mathrm{comp}}^{(\ell)}$,
and jointly train the denoiser and routers with
\begin{equation}
\mathcal{L}=\mathcal{L}_{\mathrm{diff}}
+\lambda\mathcal{L}_{\mathrm{comp}},
\label{eq:HiT-objective}
\end{equation}
where $\lambda$ controls the compression-loss weight.
Appendix~\ref{sec:compression-details} defines the
differentiable group-count estimate and gives the equivalent
original H-Net formulation.

\section{Experiments}
\subsection{Experimental Setup}
\label{sec:setup}

\paragraph{Tasks and models.}
We evaluate class-conditional generation on ImageNet at
$256\times256$ resolution~\citep{deng2009imagenet} and
text-to-image generation at $512\times512$ resolution,
training on CC12M~\citep{changpinyo2021cc12m}.
Our baselines are JiT and MiniT2I,
respectively~\citep{li2025jit,wang2026minit2i}.
Both MiniT2I and HiT use a frozen FLAN-T5-Large text encoder
with a sequence length of 256 for text-to-image
generation~\citep{chung2024flan}.
Unless otherwise stated, HiT uses two successive $2\times$
grouping stages, with patch sizes of $16$ for ImageNet and
$16$ for text-to-image generation.
For HiT's main ImageNet and text-to-image results,
including the convergence comparisons, we use
\emph{group guidance}, which contrasts denoising predictions
at different grouping capacities to steer sampling toward
the higher-capacity branch.
Its formulation and task-specific settings are detailed
in Appendix~\ref{sec:group-guidance}.

\paragraph{Training.}
We follow the training recipes of JiT for ImageNet
and MiniT2I for text-to-image
generation~\citep{li2025jit,wang2026minit2i}.
We use AdamW with $\beta=(0.9,0.95)$, zero weight
decay, and a global batch size of 1,024.
Unless otherwise stated, ImageNet ablations train HiT-B/16
for 80 epochs, following JiT-B's
training recipe~\citep{li2025jit}.
For text-to-image generation, the learning rate increases
from $10^{-6}$ to $4\times10^{-4}$ over 5,000 updates
and then remains constant, with EMA decay $0.99995$.
We retain the corresponding baseline's diffusion objective,
timestep distribution, and conditioning-dropout protocol. For HiT, we set $T_{\mathrm r}=0.05$,
$T_{\mathrm b}=0.1$, and $\lambda=0.01$ for the sum
of stage compression losses.
Thresholds are calibrated sequentially on 1,024 fixed
noisy training examples and then frozen
(Appendix~\ref{sec:training-details}).

\paragraph{Evaluation.}
We report FID~\citep{heusel2017fid} using 50,000 generated
ImageNet images and the ADM protocol~\citep{dhariwal2021adm},
with 50-step Heun sampling and a final Euler step.
For Text-to-image, we use 100-step Euler sampling.
Unless otherwise stated, ImageNet ablations report FID-50k
at epoch 80 with EMA decay $0.9999$ and classifier-free
guidance (CFG) at a fixed scale of $2.9$, applied over
$t\in(0.1,1.0)$.
Following JiT~\citep{li2025jit}, we report the forward cost
of one image in GFLOPs, counting one multiply-add as one
operation. For HiT, we report nominal costs computed using the target group count at each stage; actual costs vary with the
group counts produced for each image and diffusion timestep.

\subsection{ImageNet Generation}
\label{sec:imagenet}

\paragraph{Generation quality.}
Table~\ref{tab:imagenet} reports FID-50k on ImageNet
$256\times256$. HiT shares JiT's pixel-space formulation
and core Transformer components, making JiT our primary
baseline for assessing hierarchical modeling~\citep{li2025jit}.
HiT-XL/16 reaches 1.87 FID after 280 epochs, nearly matching
JiT-H/16's 1.86 after 600 epochs with less than half the
training updates. At 600 epochs, HiT achieves \textbf{1.48 FID}, substantially outperforming JiT-H/16 (1.86)
at a comparable parameter count.
It also surpasses PixNerd-XL/16 (2.15), PixelGen-XL/16 (1.83),
and PixelDiT-XL/16 (1.61), all of which use external visual
supervision~\citep{wang2025pixnerd,ma2026pixelgen,yu2026pixeldit},
and approaches REPA-enhanced SiT-XL/2's 1.42 FID in latent
space~\citep{yu2025repa}. These results demonstrate that learned
hierarchical processing substantially improves pixel-space
generation, narrowing the gap with strong latent models
without visual pretraining or representation alignment.

\begin{table}[t]
\centering
\setlength{\tabcolsep}{2pt}
\renewcommand{\arraystretch}{1.08}
\resizebox{\columnwidth}{!}{%
\begin{tabular}{l|cc|ccc|c}
\toprule
& \multicolumn{2}{c|}{Pre-training}
& & & & \\
\cmidrule(lr){2-3}
ImageNet $256\times256$
& Token. & Self-sup.
& Params & GFLOPs & Epochs
& FID$\downarrow$ \\
\midrule

\rowcolor[gray]{0.92}
\multicolumn{7}{l}{\textit{Latent-space diffusion}} \\
DiT-XL/2~\citep{peebles2023dit}
& SD-VAE & --
& 675+49M & 119 & 1400 & 2.27 \\
SiT-XL/2~\citep{ma2024sit}
& SD-VAE & --
& 675+49M & 119 & 1400 & 2.06 \\
REPA, SiT-XL/2~\citep{yu2025repa}
& SD-VAE & DINOv2
& 675+49M & 119 & 800 & 1.42 \\
LightningDiT-XL/2~\citep{yao2025lightningdit}
& VA-VAE & DINOv2
& 675+49M & 119 & 800 & 1.35 \\
DDT-XL/2~\citep{wang2025ddt}
& SD-VAE & DINOv2
& 675+49M & 119 & 400 & 1.26 \\
RAE, DiT$^{\mathrm{DH}}$-XL/2~\citep{zheng2025rae}
& RAE & DINOv2
& 839+415M & 146 & 800 & \textbf{1.13} \\

\midrule
\rowcolor[gray]{0.92}
\multicolumn{7}{l}{\textit{Pixel-space diffusion}} \\
ADM-G~\citep{dhariwal2021adm}
& -- & --
& 554M & 1120 & 400 & 4.59 \\
SiD, UViT/2~\citep{hoogeboom2023simplediffusion}
& -- & --
& 2B & 555 & 800 & 2.44 \\
VDM++, UViT/2~\citep{kingma2023vdm}
& -- & --
& 2B & 555 & 560 & 2.12 \\

PixelFlow-XL/4~\citep{chen2025pixelflow}
& -- & --
& 677M & 2909 & 320 & 1.98 \\
PixNerd-XL/16~\citep{wang2025pixnerd}
& -- & DINOv2
& 700M & 134 & 160 & 2.15 \\
PixelGen-XL/16~\citep{ma2026pixelgen}
& -- & DINOv2
& 676M & -- & 160 & 1.83 \\
PixelDiT-XL/16~\citep{yu2026pixeldit}
& -- & DINOv2
& 797M & 155.6 & 320 & 1.61 \\

\midrule
JiT-H/16~\citep{li2025jit}
& -- & --
& 953M & 182  & 600 & 1.86 \\

JiT-G/16~\citep{li2025jit}
& -- & --
& 2B & 383 & 320 & 1.82 \\

\midrule

\textbf{HiT-XL/16}
& -- & --
& 959.5M & 176.9 & 280 & 1.87 \\
& -- & --
& -- & -- & 600 & \textbf{1.48} \\

\bottomrule
\end{tabular}%
}
\caption{\textbf{ImageNet $256\times256$ generation.}
FID is evaluated on 50K generated images with guidance.
Parameters follow JiT's convention; GFLOPs are defined
in Sec.~\ref{sec:setup}. }
\label{tab:imagenet}
\end{table}

\subsection{Text-to-Image Generation}
\label{sec:text-to-image-results}

We next evaluate whether the learned hierarchy extends to
text-conditioned generation.
We compare HiT-B/16 with
MiniT2I~\citep{wang2026minit2i} at $512\times512$ resolution,
using the same frozen FLAN-T5-Large text encoder.
MiniT2I uses separate image and text projections and MLPs
with joint attention.
We retain these branches in HiT's outer stages and introduce
a shared hierarchy that jointly groups contextualized image
and valid caption features.
Shared Transformer blocks then process the compact group tokens.
We exclude padding and the timestep token from grouping
and the compression budget.
Appendix~\ref{sec:joint-grouping} provides further architecture details.
Our HiT--MiniT2I comparisons match training stages and numbers
of training examples, using identical data, optimizer,
and evaluation protocols.

As shown in Table~\ref{tab:text-to-image}, HiT-B/16 outperforms
MiniT2I-B/16 across all five text-to-image benchmarks,
improving the average score by 1.29 points while reducing
forward FLOPs by 31\%.
It also surpasses UniLiP-3B's average
score~\citep{tang2025unilip}, using approximately
$6\times$ fewer denoiser parameters and
$3\times$ fewer forward FLOPs.

\begin{table*}[t]
    \centering
    \small
    \setlength{\tabcolsep}{4pt}
    \renewcommand{\arraystretch}{1.08}
    \begin{tabular}{lcccccccc}
        \toprule
        & \multicolumn{2}{c}{Denoiser cost}
        & \multicolumn{6}{c}{Text-to-image benchmarks $\uparrow$} \\
        \cmidrule(lr){2-3}\cmidrule(l){4-9}
        \textbf{Model} & \textbf{Params (B)}
        & \textbf{GFLOPs}
        & \textbf{GenEval}~\cite{geneval2023}
        & \textbf{DPG}~\cite{dpg2024}
        & \textbf{PRISM}~\cite{prism2025}
        & \textbf{CoRe}~\cite{corebench2025}
        & \textbf{Spatial}~\cite{spatialgeneval2026}
        & \textbf{Avg.} \\
        \midrule
        PixArt-$\alpha$~\cite{chen2024pixartalpha}
        & 0.61 & 618.0
        & 48.00 & 71.60 & 49.56 & 24.36 & 33.70 & 45.44 \\
        PixArt-$\Sigma$~\cite{chen2024pixartsigma}
        & 0.61 & 644.3
        & 52.00 & 79.50 & 54.94 & 33.71 & 43.37 & 52.70 \\
        E-MMDiT~\cite{shen2025emmdit}
        & 0.30 & 40.2
        & 69.10 & 80.40 & 51.60 & 30.70 & 45.70 & 55.50 \\
        SANA-0.6B~\cite{xie2024sana}
        & 0.59 & 157.8
        & 65.80 & 83.30 & 56.40 & 39.80 & 47.40 & 58.54 \\
        PixNerd-XXL/16~\cite{wang2025pixnerd}
        & 1.17 & 768.9
        & 73.00 & 80.90 & 52.45 & 36.00 & 49.28 & 58.33 \\
        PixelDiT-T2I~\cite{yu2026pixeldit}
        & 1.31 & 716.3
        & 78.00 & 83.70 & -- & -- & -- & -- \\
        DeCo-XXL/16~\cite{ma2025deco}
        & 1.10 & 725.3
        & 82.00 & 82.00 & 52.60 & 34.80 & 49.80 & 60.24 \\
        TiM-T2I~\cite{wang2025tim}
        & 0.87 & 167.1
        & 82.80 & 83.20 & 51.40 & 36.70 & 48.20 & 60.46 \\
        MiniT2I-B/16~\cite{wang2026minit2i}
        & 0.26 & 201.6
        & 87.50 & \underline{84.10} & 55.60
        & 44.00 & \underline{52.00} & 64.64 \\
        UniLiP-3B~\cite{tang2025unilip}
        & 1.60 & 408.0
        & \textbf{90.30} & 83.40 & \underline{58.50}
        & \underline{45.00} & 51.60 & \underline{65.76} \\
        \midrule
        \textbf{HiT-B/16}
        & 0.25 & 139.4
        & \underline{87.78} & \textbf{85.33} & \textbf{58.62}
        & \textbf{45.22} & \textbf{52.68} & \textbf{65.93} \\
        \bottomrule
    \end{tabular}
    \caption{
        \textbf{Text-to-image generation at $512\times512$.}
        Best scores are bold; second-best scores are underlined.
    }
    \label{tab:text-to-image}
\end{table*}

\subsection{Analysis of Learned Groups}
\label{sec:group-analysis}

\begin{figure*}[t]
    \centering
    \includegraphics[width=\textwidth]{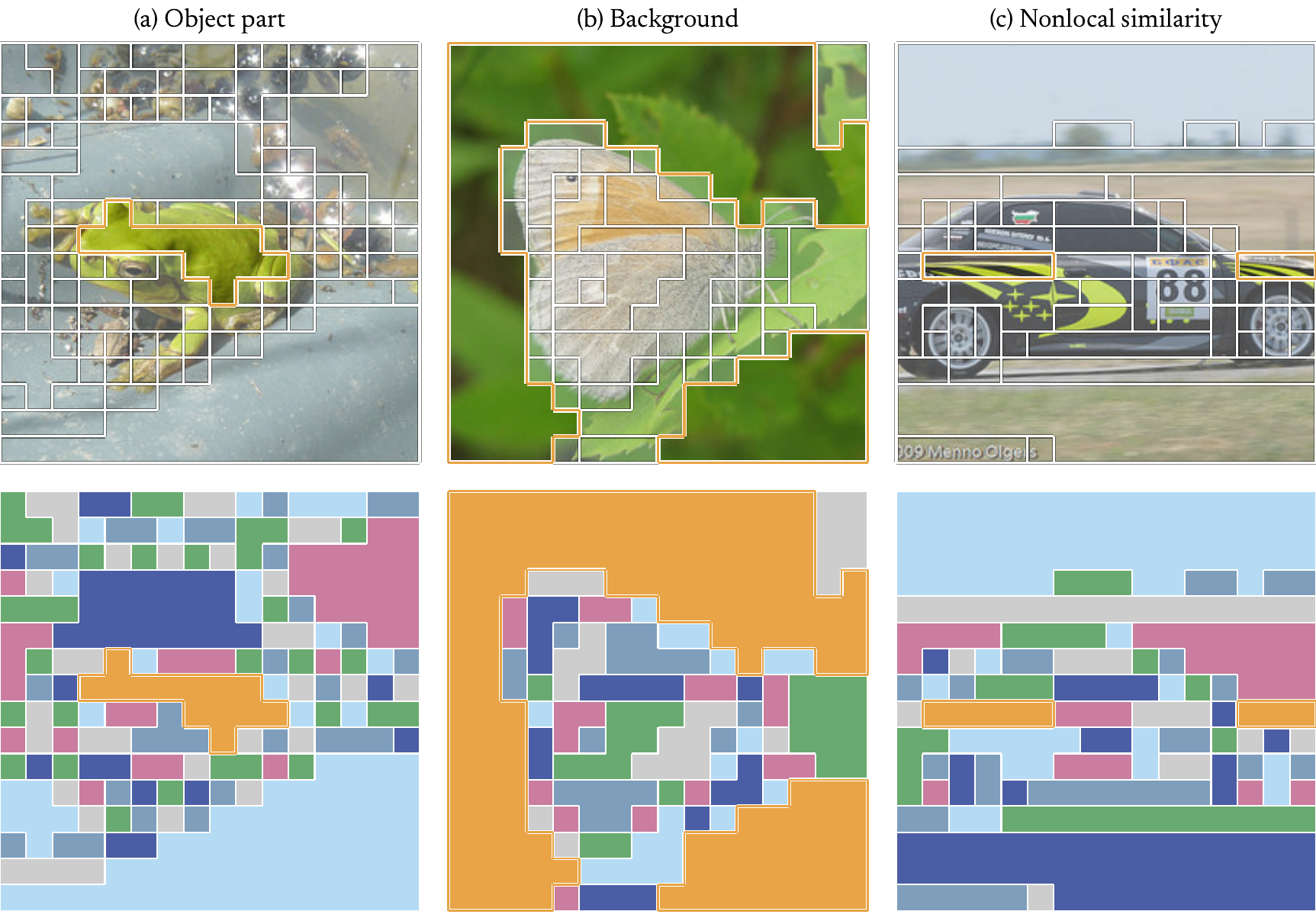}
    \caption{
        \textbf{Learned visual organization of HiT.}
        Groups are computed based on noisy inputs at $t=0.75$ from ImageNet.
        The top row overlays group boundaries on clean images;
        the bottom row shows the corresponding hard group maps.
        Yellow highlights one group in each column:
        an object region, background foliage, or spatially
        separated car markings.
        Other colors distinguish adjacent groups and may
        repeat for distant groups.
    }
    \label{fig:visual-groups}
\end{figure*}

\paragraph{Learned visual organization.}
Figure~\ref{fig:visual-groups} shows the groups of HiT-B/16 at $t=0.75$ on different images from ImageNet.
Some groups follow coherent object regions, while others
collect repeated background structures, such as foliage,
into a shared representation.
The highlighted car markings further show that spatially
disconnected patches can belong to the same group.
Together, these examples show that HiT learns computational
units reflecting object structure and appearance similarity,
extending beyond local spatial neighborhoods.
This organization emerges from generative training without
segmentation annotations.

 \paragraph{Learned image--text associations.}
HiT connects caption words to corresponding image regions
and image patches to relevant words, without word--region
annotations.
We provide visualizations on COCO image--caption
pairs~\citep{lin2014coco} in Fig.~\ref{fig:teaser-routing} in
Appendix~\ref{sec:additional-visualizations}
, with measurement definitions
in Appendix~\ref{sec:routing-visualization-details}.

\subsection{Ablation Studies}
\label{sec:ablations}

\paragraph{Controlled setting.}
Our reference is an 18-block HiT-B/16 with width 768
and two successive $2\times$ grouping stages.
Its $3/3/6/3/3$ layout specifies the block counts in the
full-resolution encoder, intermediate encoder, compact core,
intermediate decoder, and full-resolution decoder, respectively.
All five-entry layouts follow this order.
We use linear patch embedding, eight class-context tokens
and one time token from the first block, and no adaLN.
Other settings follow Sec.~\ref{sec:setup}.
Unless stated otherwise, each ablation changes only the
indicated component.
Gray rows mark the reference configuration. We present more ablation in Appendix \ref{sec:additional-ablations}.

\paragraph{Grouping and Ungrouping.}
We ablate how grouping constructs compact group features and how
ungrouping restores features to image positions
(Table~\ref{tab:ablation-readers}).
All variants use the same hard-grouping rule.
At each level, the router produces a partition
$\{G_g\}_{g=1}^{K}$ and learned soft weights $A_{ig}$,
where $\sum_g A_{ig}=1$.

For grouping, the default aggregates all input tokens using
these learned weights:
\begin{equation}
    z_g =
    \frac{\sum_{i=1}^{N} A_{ig}h_i}
         {\sum_{i=1}^{N} A_{ig}},
\end{equation}
where $h_i$ is an input token feature.
We compare this with two alternatives that pool only
tokens within the hard group $G_g$.
\emph{Uniform} weighting averages its members equally;
\emph{learned} weighting retains $A_{ig}$ but renormalizes
over those members.
Both alternatives retain the default soft ungrouping.

For ungrouping, the default restores each image position through
a soft mixture of processed group features:
\begin{equation}
    r_i = \sum_{g=1}^{K} A_{ig}\widetilde{z}_g.
\end{equation}
The weights computed before compact processing are reused
for readback.
We replace this mixture with \emph{hard lookup},
$r_i=\widetilde{z}_{g(i)}$, where $g(i)$ is the token's
assigned hard group, while retaining the default grouping.
Each modification applies to both hierarchy levels,
preserves the stage skips, and uses identical operations
during training and inference.

Restricting aggregation to hard-group members worsens FID
from 6.61 to 10.16--10.32, while hard readback increases it
to 7.97. These results favor soft information exchange
across hard-group boundaries during both aggregation
and readback.

\begin{table}[t]
    \centering
    \small
    \setlength{\tabcolsep}{3pt}
    \renewcommand{\arraystretch}{1.10}
    \begin{tabular}{lllc}
        \toprule
        \multicolumn{2}{c}{\textbf{Group}}
        & \textbf{Ungroup}
        & \textbf{FID$\downarrow$} \\
        \cmidrule(lr){1-2}
        \textbf{Tokens}
        & \textbf{Weights}
        & & \\
        \midrule
        Within hard group & Uniform
        & Soft mixture & 10.32 \\
        Within hard group & Learned
        & Soft mixture & 10.16 \\
        All input tokens & Learned
        & Hard lookup & 7.97 \\
        \midrule
        \rowcolor{gray!12}
        All input tokens & Learned
        & Soft mixture & \textbf{6.61} \\
        \bottomrule
    \end{tabular}
    \caption{
        \textbf{Grouping and ungrouping.}
        We vary which tokens each group pools from and how ungrouping restores
        token features.}
    \label{tab:ablation-readers}
\end{table}

\paragraph{Compression hierarchy.}
We study both compression strength and where it occurs
(Table~\ref{tab:ablation-compression}).
Two-stage variants use layout $3/3/6/3/3$, while the
single-stage $4\times$ variant uses $3/12/3$.
All variants retain 18 blocks, the same width and conditioning,
but differ in computational cost.
Token counts are target averages; actual group counts remain
input dependent.
The $2\times4$ and $4\times2$ variants both target 32 core
tokens, allowing us to compare earlier and later compression.

The default $2\times2$ hierarchy achieves the lowest FID
of 6.61, while $2\times4$ remains close at 6.89 with
modestly lower FLOPs.
Despite the same core token target, $4\times2$ performs
substantially worse at 9.54, suggesting that retaining more
tokens for intermediate processing is beneficial.
Stronger $4\times4$ compression and the single-stage
$4\times$ variant also reduce generation quality.
These results favor moderate, progressive compression,
and we retain $2\times2$ as our default.
\begin{table}[t]
    \centering
    \small
    \setlength{\tabcolsep}{3pt}
    \renewcommand{\arraystretch}{1.10}
       \begin{tabular}{lcccc}
        \toprule
        \textbf{Model}
        & \textbf{Ratio}
        & \textbf{Image tokens}
        & \textbf{GFLOPs}
        & \textbf{FID$\downarrow$} \\
        \midrule
        HiT-B/16
        & $4\times$
        & $256\rightarrow64$
        & 19.1 & 9.59 \\
        HiT-B/16
        & $2\times4$
        & $256\rightarrow128\rightarrow32$
        & 20.8 &  6.89\\
        HiT-B/16
        & $4\times2$
        & $256\rightarrow64\rightarrow32$
        & 17.8 & 9.54\\
        HiT-B/16
        & $4\times4$
        & $256\rightarrow64\rightarrow16$
        & 17.1 & 10.78\\

        \midrule
        \rowcolor{gray!12}
        \textbf{HiT-B/16}
        & $\mathbf{2\times2}$
        & $256\rightarrow128\rightarrow64$
        & \textbf{22.2}
        & \textbf{6.61} \\
        \bottomrule
    \end{tabular}
    \caption{
        \textbf{Compression hierarchy.}
        We vary the compression ratio of each grouping stage,
        using one or two stages and 18 blocks.}
    \label{tab:ablation-compression}
\end{table}

\paragraph{Block allocation.}
We redistribute 18 Transformer blocks across HiT-B/16,
preserving encoder--decoder symmetry, width 768, and
$2\times2$ grouping.
We vary the outer and intermediate depths independently
between two and three blocks, assigning the remaining
blocks to the compact core.
All layouts retain approximately $133.9$M parameters,
but their FLOPs differ because blocks operate at different
token resolutions.

Table~\ref{tab:ablation-allocation} shows that moving blocks
from the compact core to the outer and intermediate stages
improves FID from 8.48 to 6.61, while increasing forward
cost from 18.3 to 22.2 GFLOPs.
With eight core blocks, allocating more depth to full-resolution
processing ($3/2/8/2/3$) achieves better FID than allocating it
to intermediate processing ($2/3/8/3/2$): 6.81 versus 7.60.
These results highlight the value of processing at finer
resolutions alongside the compact core.
We retain $3/3/6/3/3$ as the default, which achieves the
lowest FID among the tested layouts.

\begin{table}[t]
    \centering
    \small
    \setlength{\tabcolsep}{5pt}
    \renewcommand{\arraystretch}{1.10}
    \begin{tabular}{lcccc}
        \toprule
        \textbf{Model}
        & \textbf{Block layout}
        & \textbf{GFLOPs}
        & \textbf{FID$\downarrow$} \\
        \midrule
        HiT-B/16
        & $2/2/10/2/2$
        & 18.3 & 8.48 \\
        HiT-B/16
        & $2/3/8/3/2$
        & 19.3 & 7.60\\
        HiT-B/16
        & $3/2/8/2/3$
        & 21.3 & 6.81\\

        \midrule
        \rowcolor{gray!12}
        \textbf{HiT-B/16}
        & $\mathbf{3/3/6/3/3}$
        & \textbf{22.2}
        & \textbf{6.61} \\
        \bottomrule
    \end{tabular}
    \caption{
        \textbf{Block allocation.}
        We redistribute the same 18 Transformer blocks
        across hierarchy levels.
}
    \label{tab:ablation-allocation}
\end{table}

\paragraph{Conditioning and capacity.}
We investigate whether parameters used for adaLN are better
allocated to Transformer depth
(Table~\ref{tab:ablation-conditioning}).
The adaLN variants modulate the blocks and output head
with timestep/class embeddings and introduce class-context
tokens after the fourth block.
The token-conditioned variants remove adaLN and introduce
class and time tokens before the first block, changing
both the conditioning mechanism and token placement.

For 12-block HiT-B/16 with layout $2/2/4/2/2$,
token conditioning reduces parameters from $135.0$M to
$91.3$M while improving FID from $12.27$ to $11.55$.
Increasing depth to 18 blocks with layout $3/3/6/3/3$
restores the parameter count to $133.9$M and further
improves FID to $6.61$, with other settings unchanged.

The analogous reallocation in JiT-B instead worsens FID
from $10.37$ to $12.85$ at comparable parameter counts.
We hypothesize that grouping facilitates interactions
between conditioning tokens and compact image representations,
helping HiT exploit the additional depth.
These results favor combining token conditioning with
deeper hierarchical processing to use HiT's parameter
budget effectively.

\begin{table}[t]
    \centering
    \small
    \setlength{\tabcolsep}{3pt}
    \renewcommand{\arraystretch}{1.10}
    \begin{tabular}{lccccc}
        \toprule
        \textbf{Model}
        & \textbf{Cond.}
        & \textbf{Depth}
        & \textbf{Params}
        & \textbf{GFLOPs}
        & \textbf{FID$\downarrow$} \\
        \midrule
        JiT-B
        & adaLN & 12 & 131.1M & 25.2 & 10.37 \\
        JiT-B
        & In-context & 18 & 129.9M & 39.3 & 12.85 \\

        \midrule
        HiT-B/16
        & adaLN & 12 & 135.0M & 14.9 & 12.27 \\
        HiT-B/16
        & In-context & 12 & 91.3M & 15.2 & 11.55 \\

        \midrule
        \rowcolor{gray!12}
        \textbf{HiT-B/16}
        & \textbf{In-context}
        & \textbf{18}
        & \textbf{133.9M}
        & \textbf{22.2}
        & \textbf{6.61} \\
        \bottomrule
    \end{tabular}
    \caption{
        \textbf{Trading adaLN parameters for depth.}
        We replace adaLN with in-context conditioning and
        spend the saved parameters on more blocks.}
    \label{tab:ablation-conditioning}
\end{table}

\paragraph{Learned versus fixed grouping.}
We compare learned grouping with a fixed spatial hierarchy
using the same HiT-B/16 backbone.
The fixed control pairs adjacent patches horizontally at
stage 1 and adjacent groups vertically at stage 2,
producing exactly $256\rightarrow128\rightarrow64$ tokens.
It retains learned query/key projections and input-dependent
soft pooling and ungrouping, but omits tree construction
and the compression loss.
The learned variant instead adapts both partitions and group
counts, targeting the same average token populations.

Table~\ref{tab:ablation-fixed-grouping} shows that learned
grouping improves FID from 7.11 to 6.61 with identical
parameter counts (133.9M) and nearly matched forward costs
(22.1 versus 22.2 GFLOPs).
This improvement supports learning the partitions and group
counts jointly, beyond learning soft feature exchange over
a fixed spatial hierarchy.

\begin{table}[t]
    \centering
    \small
    \setlength{\tabcolsep}{3pt}
    \renewcommand{\arraystretch}{1.10}
    \begin{tabular}{lcccc}
        \toprule
        \textbf{Model}
        & \textbf{Grouping}
        & \textbf{Counts}
        & \textbf{GFLOPs}
        & \textbf{FID$\downarrow$} \\
        \midrule
        HiT-B/16
        & Spatial
        & Fixed
        & 22.1
        & 7.11 \\

        \midrule
        \rowcolor{gray!12}
        \textbf{HiT-B/16}
        & \textbf{Learned}
        & \textbf{Adaptive}
        & \textbf{22.2}
        & \textbf{6.61} \\
        \bottomrule
    \end{tabular}
    \caption{
        \textbf{Learned versus fixed grouping.}
        We compare learned, adaptive groups with fixed
        groups of neighboring patches.}
    \label{tab:ablation-fixed-grouping}
\end{table}

\paragraph{Model architecture and scaling.}
We examine whether HiT's improvements persist as model
capacity increases.
HiT-B, HiT-L, and HiT-XL approximately match the parameter
counts of JiT-B, JiT-L, and JiT-H,
respectively~\citep{li2025jit}.
Table~\ref{tab:model-architectures} reports their architectures,
computational costs, and generation quality.
HiT consistently improves generation quality across
model scales.
At the B, L, and XL scales, HiT achieves lower FID than the
corresponding JiT model while requiring fewer forward FLOPs.
All comparisons use 80 training epochs and CFG 2.9.

\begin{table}[t]
    \centering
    \small
    \setlength{\tabcolsep}{2pt}
    \renewcommand{\arraystretch}{1.10}
    \begin{tabular*}{\columnwidth}{
        @{}l@{\extracolsep{\fill}}ccccc
    }
        \toprule
        \textbf{Model}
        & \textbf{Width}
        & \textbf{Layout}
        & \textbf{Params}
        & \textbf{GFLOPs}
        & \textbf{FID$\downarrow$} \\
        \midrule
        JiT-B
        & 768 & 12
        & 131.1M & 25.2 & 10.37 \\
        \HiTrowshade
        \textbf{HiT-B}
        & 768 & $3/3/6/3/3$
        & 133.9M & 22.2 & \textbf{6.61} \\

        \midrule
        JiT-L
        & 1024 & 24
        & 458.9M & 87.9 & 4.85 \\
        \HiTrowshade
        \textbf{HiT-L}
        & 1024 & $3/3/24/3/3$
        & 463.6M & 55.6 & \textbf{4.16} \\

        \midrule
        JiT-H
        & 1280 & 32
        & 952.5M & 181.9 & 4.39 \\
        \HiTrowshade
        \textbf{HiT-XL}
        & 1280 & $10/10/8/10/10$
        & 959.5M & 176.9 & \textbf{3.64} \\
             \bottomrule

    \end{tabular*}
    \caption{
        \textbf{Scaling.}
        We compare HiT and JiT at three sizes with similar
        parameter counts, after 80 epochs.
    }
    \label{tab:model-architectures}
\end{table}

\section{Conclusion}

We introduced HiT, which learns a visual hierarchy within
a single end-to-end pixel-space generator.
By grouping related features into compact units, HiT
organizes visual information while reducing the cost
of deeper processing.
The learned groups capture coherent object structure and
image--text associations without region annotations.
HiT achieves 1.48 FID on ImageNet, reaches JiT's quality
with substantially less training compute, and outperforms
MiniT2I across five text-to-image benchmarks.
These results support learned visual hierarchy as a
foundation for strong, efficient pixel-space generation
without separate visual pretraining.
{
    \small
    \bibliographystyle{ieeenat_fullname}
    \bibliography{main}

@String(CVPR= {IEEE Conf. Comput. Vis. Pattern Recog.})

@String(ICCV= {Int. Conf. Comput. Vis.})

@String(CVPRW= {IEEE Conf. Comput. Vis. Pattern Recog. Worksh.})

@String(CVPR  = {CVPR})

@String(ICCV  = {ICCV})

@String(CVPRW= {CVPRW})

@article{bolya2023tome,
  title={Token merging: Your vit but faster},
  author={Bolya, Daniel and Fu, Cheng-Yang and Dai, Xiaoliang and Zhang, Peizhao and Feichtenhofer, Christoph and Hoffman, Judy},
  journal={arXiv preprint arXiv:2210.09461},
  year={2022}
}

@article{rombach2022ldm,
  title={High-Resolution Image Synthesis with Latent Diffusion Models},
  author={Rombach, Robin},
  journal={arXiv preprint arXiv:2112.10752},
  year={2022}
}

@inproceedings{peebles2023dit,
  title={Scalable diffusion models with transformers},
  author={Peebles, William and Xie, Saining},
  booktitle={2023 IEEE/CVF International Conference on Computer Vision (ICCV)},
  pages={4172--4182},
  year={2023},
  organization={IEEE}
}

@inproceedings{zheng2025rae,
  title={Diffusion transformers with representation autoencoders},
  author={Zheng, Boyang and Ma, Nanye and Tong, Shengbang and Xie, Saining},
  booktitle={International Conference on Learning Representations},
  volume={2026},
  pages={35791--35820},
  year={2026}
}

@article{yu2025repa,
  title={Representation alignment for generation: Training diffusion transformers is easier than you think},
  author={Yu, Sihyun and Kwak, Sangkyung and Jang, Huiwon and Jeong, Jongheon and Huang, Jonathan and Shin, Jinwoo and Xie, Saining},
  journal={arXiv preprint arXiv:2410.06940},
  year={2024}
}

@inproceedings{hoogeboom2023simplediffusion,
  title={simple diffusion: End-to-end diffusion for high resolution images},
  author={Hoogeboom, Emiel and Heek, Jonathan and Salimans, Tim},
  booktitle={International Conference on Machine Learning},
  pages={13213--13232},
  year={2023},
  organization={PMLR}
}

@inproceedings{hoogeboom2025sid2,
  title={Simpler diffusion: 1.5 fid on imagenet512 with pixel-space diffusion},
  author={Hoogeboom, Emiel and Mensink, Thomas and Heek, Jonathan and Lamerigts, Kay and Gao, Ruiqi and Salimans, Tim},
  booktitle={2025 IEEE/CVF Conference on Computer Vision and Pattern Recognition (CVPR)},
  pages={18062--18071},
  year={2025},
  organization={IEEE}
}

@article{chen2025pixelflow,
  title={Pixelflow: Pixel-space generative models with flow},
  author={Chen, Shoufa and Ge, Chongjian and Zhang, Shilong and Sun, Peize and Luo, Ping},
  journal={arXiv preprint arXiv:2504.07963},
  year={2025}
}

@inproceedings{yu2026pixeldit,
  title={Pixeldit: Pixel diffusion transformers for image generation},
  author={Yu, Yongsheng and Xiong, Wei and Nie, Weili and Sheng, Yichen and Liu, Shiqiu and Luo, Jiebo},
  booktitle={Proceedings of the IEEE/CVF Conference on Computer Vision and Pattern Recognition},
  pages={14273--14282},
  year={2026}
}

@inproceedings{li2025jit,
  title={Back to basics: Let denoising generative models denoise},
  author={Li, Tianhong and He, Kaiming},
  booktitle={Proceedings of the IEEE/CVF Conference on Computer Vision and Pattern Recognition},
  pages={36115--36125},
  year={2026}
}

@inproceedings{hwang2025hnet,
  title={Dynamic chunking for end-to-end hierarchical sequence modeling},
  author={Hwang, Sukjun and Wang, Brandon and Gu, Albert},
  booktitle={International Conference on Learning Representations},
  volume={2026},
  pages={149273--149313},
  year={2026}
}

@misc{wang2026minit2i,
  title  = {MiniT2I: A Minimalist Baseline for Text-to-Image Generation},
  author = {Wang, Xianbang and Zhao, Hanhong and Lu, Yiyang and Zhou, Kangyang and Ma, Linrui and He, Kaiming},
  year   = {2026},
  url    = {https://peppaking8.github.io/minit2i/}
}

@article{cartesia2026hnet,
  author  = {{Cartesia Research}},
  title   = {From Tokens to Bytes: Scaling Hierarchical Networks},
  journal = {Cartesia Blog},
  year    = {2026},
  month   = {October},
  note    = {https://www.cartesia.ai/blog/from-tokens-to-bytes-part-1}
}

@inproceedings{pagnoni2025blt,
  title={Byte latent transformer: Patches scale better than tokens},
  author={Pagnoni, Artidoro and Pasunuru, Ramakanth and Rodriguez, Pedro and Nguyen, John and Muller, Benjamin and Li, Margaret and Zhou, Chunting and Yu, Lili and Weston, Jason E and Zettlemoyer, Luke and others},
  booktitle={Proceedings of the 63rd Annual Meeting of the Association for Computational Linguistics (Volume 1: Long Papers)},
  pages={9238--9258},
  year={2025}
}

@inproceedings{bolya2023tomesd,
  title={Token merging for fast stable diffusion},
  author={Bolya, Daniel and Hoffman, Judy},
  booktitle={2023 IEEE/CVF conference on computer vision and pattern recognition workshops (CVPRW)},
  pages={4599--4603},
  year={2023},
  organization={IEEE}
}

@inproceedings{xu2022groupvit,
  title={Groupvit: Semantic segmentation emerges from text supervision},
  author={Xu, Jiarui and De Mello, Shalini and Liu, Sifei and Byeon, Wonmin and Breuel, Thomas and Kautz, Jan and Wang, Xiaolong},
  booktitle={2022 IEEE/CVF Conference on Computer Vision and Pattern Recognition (CVPR)},
  pages={18113--18123},
  year={2022},
  organization={IEEE}
}

@article{zhang2019rmsnorm,
  title={Root mean square layer normalization},
  author={Zhang, Biao and Sennrich, Rico},
  journal={Advances in neural information processing systems},
  volume={32},
  year={2019}
}

@inproceedings{henry2020qknorm,
  title={Query-key normalization for transformers},
  author={Henry, Alex and Dachapally, Prudhvi Raj and Pawar, Shubham Shantaram and Chen, Yuxuan},
  booktitle={Findings of the Association for Computational Linguistics: EMNLP 2020},
  pages={4246--4253},
  year={2020}
}

@article{shazeer2020glu,
  title={Glu variants improve transformer},
  author={Shazeer, Noam},
  journal={arXiv preprint arXiv:2002.05202},
  year={2020}
}

@article{su2024roformer,
  title={Roformer: Enhanced transformer with rotary position embedding},
  author={Su, Jianlin and Ahmed, Murtadha and Lu, Yu and Pan, Shengfeng and Bo, Wen and Liu, Yunfeng},
  journal={Neurocomputing},
  volume={568},
  pages={127063},
  year={2024},
  publisher={Elsevier}
}

@inproceedings{deng2009imagenet,
  title={Imagenet: A large-scale hierarchical image database},
  author={Deng, Jia and Dong, Wei and Socher, Richard and Li, Li-Jia and Li, Kai and Fei-Fei, Li},
  booktitle={2009 IEEE conference on computer vision and pattern recognition},
  pages={248--255},
  year={2009},
  organization={Ieee}
}

@inproceedings{changpinyo2021cc12m,
  title={Conceptual 12m: Pushing web-scale image-text pre-training to recognize long-tail visual concepts},
  author={Changpinyo, Soravit and Sharma, Piyush and Ding, Nan and Soricut, Radu},
  booktitle={2021 IEEE/CVF Conference on Computer Vision and Pattern Recognition (CVPR)},
  pages={3557--3567},
  year={2021},
  organization={IEEE}
}

@article{chung2024flan,
  title={Scaling instruction-finetuned language models},
  author={Chung, Hyung Won and Hou, Le and Longpre, Shayne and Zoph, Barret and Tay, Yi and Fedus, William and Li, Yunxuan and Wang, Xuezhi and Dehghani, Mostafa and Brahma, Siddhartha and others},
  journal={Journal of machine learning research},
  volume={25},
  number={70},
  pages={1--53},
  year={2024}
}

@article{heusel2017fid,
  title={Gans trained by a two time-scale update rule converge to a local nash equilibrium},
  author={Heusel, Martin and Ramsauer, Hubert and Unterthiner, Thomas and Nessler, Bernhard and Hochreiter, Sepp},
  journal={Advances in neural information processing systems},
  volume={30},
  year={2017}
}

@article{dhariwal2021adm,
  title={Diffusion models beat gans on image synthesis},
  author={Dhariwal, Prafulla and Nichol, Alexander},
  journal={Advances in neural information processing systems},
  volume={34},
  pages={8780--8794},
  year={2021}
}

@inproceedings{lin2014coco,
  title={Microsoft coco: Common objects in context},
  author={Lin, Tsung-Yi and Maire, Michael and Belongie, Serge and Hays, James and Perona, Pietro and Ramanan, Deva and Doll{\'a}r, Piotr and Zitnick, C Lawrence},
  booktitle={European conference on computer vision},
  pages={740--755},
  year={2014},
  organization={Springer}
}

@inproceedings{yao2025lightningdit,
  title={Reconstruction vs. generation: Taming optimization dilemma in latent diffusion models},
  author={Yao, Jingfeng and Yang, Bin and Wang, Xinggang},
  booktitle={2025 IEEE/CVF Conference on Computer Vision and Pattern Recognition (CVPR)},
  pages={15703--15712},
  year={2025},
  organization={IEEE}
}

@inproceedings{wang2025ddt,
  title={Ddt: Decoupled diffusion transformer},
  author={Wang, Shuai and Tian, Zhi and Huang, Weilin and Wang, Limin},
  booktitle={Proceedings of the IEEE/CVF conference on computer vision and pattern recognition},
  pages={40633--40642},
  year={2026}
}

@article{kingma2023vdm,
  title={Understanding diffusion objectives as the elbo with simple data augmentation},
  author={Kingma, Diederik and Gao, Ruiqi},
  journal={Advances in Neural Information Processing Systems},
  volume={36},
  pages={65484--65516},
  year={2023}
}

@inproceedings{wang2025pixnerd,
  title={Pixnerd: Pixel neural field diffusion},
  author={Wang, Shuai and Gao, Ziteng and Zhu, Chenhui and Huang, Weilin and Wang, Limin},
  booktitle={International Conference on Learning Representations},
  volume={2026},
  pages={43559--43580},
  year={2026}
}

@inproceedings{ma2024sit,
  title={Sit: Exploring flow and diffusion-based generative models with scalable interpolant transformers},
  author={Ma, Nanye and Goldstein, Mark and Albergo, Michael S and Boffi, Nicholas M and Vanden-Eijnden, Eric and Xie, Saining},
  booktitle={European Conference on Computer Vision},
  pages={23--40},
  year={2024},
  organization={Springer}
}

@misc{dieleman2025latents,
  author = {Dieleman, Sander},
  title = {Generative modelling in latent space},
  url = {https://sander.ai/2025/04/15/latents.html},
  year = {2025}
}

@inproceedings{bao2023uvit,
  title={All are worth words: A vit backbone for diffusion models},
  author={Bao, Fan and Nie, Shen and Xue, Kaiwen and Cao, Yue and Li, Chongxuan and Su, Hang and Zhu, Jun},
  booktitle={2023 IEEE/CVF Conference on Computer Vision and Pattern Recognition (CVPR)},
  pages={22669--22679},
  year={2023},
  organization={IEEE}
}

@article{ma2026pixelgen,
  title={PixelGen: Improving pixel diffusion with perceptual supervision},
  author={Ma, Zehong and Xu, Ruihan and Zhang, Shiliang},
  journal={arXiv preprint arXiv:2602.02493},
  year={2026}
}

@article{geneval2023,
  title={Geneval: An object-focused framework for evaluating text-to-image alignment},
  author={Ghosh, Dhruba and Hajishirzi, Hannaneh and Schmidt, Ludwig},
  journal={Advances in Neural Information Processing Systems},
  volume={36},
  pages={52132--52152},
  year={2023}
}

@article{dpg2024,
  title={Ella: Equip diffusion models with llm for enhanced semantic alignment},
  author={Hu, Xiwei and Wang, Rui and Fang, Yixiao and Fu, Bin and Cheng, Pei and Yu, Gang},
  journal={arXiv preprint arXiv:2403.05135},
  year={2024}
}

@inproceedings{prism2025,
  title={Flux-reason-6m \& prism-bench: A million-scale text-to-image reasoning dataset and comprehensive benchmark},
  author={Fang, Rongyao and Yu, Aldrich and Duan, Chengqi and Huang, Linjiang and Bai, Shuai and Cai, Yuxuan and Wang, Kun and Liu, Si and Liu, Xihui and Li, Hongsheng},
  booktitle={International Conference on Learning Representations},
  volume={2026},
  pages={63157--63186},
  year={2026}
}

@inproceedings{corebench2025,
  title={Easier Painting Than Thinking: Can Text-to-Image Models Set the Stage, but Not Direct the Play?},
  author={Li, Ouxiang and Wang, Yuan and Hu, Xinting and Huang, Huijuan and Chen, Rui and Ou, Jiarong and Tao, Xin and Wan, Pengfei and Qi, Xiaojuan and Feng, Fuli},
  booktitle={International Conference on Learning Representations},
  volume={2026},
  pages={86729--86758},
  year={2026}
}

@article{spatialgeneval2026,
  title={Everything in its place: Benchmarking spatial intelligence of text-to-image models},
  author={Wang, Zengbin and Hu, Xuecai and Wang, Yong and Xiong, Feng and Zhang, Man and Chu, Xiangxiang},
  journal={arXiv preprint arXiv:2601.20354},
  year={2026}
}

@article{shen2025emmdit,
  title={E-mmdit: Revisiting multimodal diffusion transformer design for fast image synthesis under limited resources},
  author={Shen, Tong and Yu, Jingai and Zhou, Dong and Li, Dong and Barsoum, Emad},
  journal={arXiv preprint arXiv:2510.27135},
  year={2025}
}

@article{xie2024sana,
  title={Sana: Efficient high-resolution image synthesis with linear diffusion transformers},
  author={Xie, Enze and Chen, Junsong and Chen, Junyu and Cai, Han and Tang, Haotian and Lin, Yujun and Zhang, Zhekai and Li, Muyang and Zhu, Ligeng and Lu, Yao and others},
  journal={arXiv preprint arXiv:2410.10629},
  year={2024}
}

@inproceedings{ma2025deco,
  title={Deco: Frequency-decoupled pixel diffusion for end-to-end image generation},
  author={Ma, Zehong and Wei, Longhui and Wang, Shuai and Zhang, Shiliang and Tian, Qi},
  booktitle={Proceedings of the IEEE/CVF Conference on Computer Vision and Pattern Recognition},
  pages={43600--43610},
  year={2026}
}

@inproceedings{wang2025tim,
  title={Transition models: Rethinking the generative learning objective},
  author={Wang, Zidong and Zhang, Yiyuan and Yue, Xiaoyu and Yue, Xiangyu and Li, Yangguang and Ouyang, Wanli and Bai, Lei},
  booktitle={Proceedings of the IEEE/CVF Conference on Computer Vision and Pattern Recognition},
  pages={29178--29189},
  year={2026}
}

@inproceedings{tang2025unilip,
  title={Unilip: Adapting clip for unified multimodal understanding, generation and editing},
  author={Tang, Hao and Xie, Chen-Wei and Bao, Xiaoyi and Weng, Tingyu and Li, Pandeng and Zheng, Yun and Wang, Liwei},
  booktitle={International Conference on Learning Representations},
  volume={2026},
  pages={60440--60465},
  year={2026}
}

@inproceedings{chen2024pixartalpha,
  title={PixArt-$\alpha$: Fast Training of Diffusion Transformer for Photorealistic Text-to-Image Synthesis},
  author={Chen, Junsong and Yu, Jincheng and Ge, Chongjian and Yao, Lewei and Xie, Enze and Wu, Yue and Wang, Zhongdao and Kwok, James and Luo, Ping and Lu, Huchuan and Li, Zhenguo},
  booktitle={International Conference on Learning Representations},
  year={2024}
}

@inproceedings{chen2024pixartsigma,
  title={PixArt-$\Sigma$: Weak-to-Strong Training of Diffusion Transformer for 4K Text-to-Image Generation},
  author={Chen, Junsong and Ge, Chongjian and Xie, Enze and Wu, Yue and Yao, Lewei and Ren, Xiaozhe and Wang, Zhongdao and Luo, Ping and Lu, Huchuan and Li, Zhenguo},
  booktitle={European Conference on Computer Vision},
  year={2024}
}

@article{krause2026sparseguidance,
  title={Guiding Token-Sparse Diffusion Models},
  author={Krause, Felix and Baumann, Stefan Andreas and Schusterbauer, Johannes and Grebenkova, Olga and Gui, Ming and Hu, Vincent Tao and Ommer, Bj{\"o}rn},
  journal={arXiv preprint arXiv:2601.01608},
  year={2026}
}
}

\clearpage
\ifdefined\HiTArxivVersion
    \section*{Appendix}
\else
    \setcounter{page}{1}
    \maketitlesupplementary
\fi

\appendix
\section{Additional Method Details}

\subsection{Grouping Details}
\label{sec:grouping-details}
The affinity between groups $U$ and $V$ is the mean of all
pairwise token affinities:
\begin{equation}
s(U,V)
= \frac{1}{|U|\,|V|}
  \sum_{i \in U}\sum_{j \in V} S_{ij}.
\label{eq:HiT-linkage}
\end{equation}
For token $i$ and hard group $G_g$, the soft assignment is
\begin{equation}
e_{ig}
= \frac{1}{|G_g|}
  \sum_{j \in G_g} S_{ij},
\qquad
A_{ig}
= \frac{\exp(e_{ig}/T_{\mathrm{r}})}
       {\sum_{g'=1}^{K}\exp(e_{ig'}/T_{\mathrm{r}})},
\label{eq:HiT-assignment}
\end{equation}
where $T_{\mathrm{r}}$ controls the sharpness of the weights.
Each row of $A$ sums to one.

The pooling matrix has entries $P_{ig}=A_{ig}/\sum_j A_{jg}$.
Thus, each compact feature is a weighted average of the input features,
as given by Eq.~\eqref{eq:HiT-stage}.
Ungrouping produces $R=A\mathcal{M}(Z)$.
It restores token positions but does not invert the lossy pooling operation.
The soft weights allow every input token to contribute to each group.

\subsection{Compression Loss Details}
\label{sec:compression-details}
For one stage, the ratio loss~\citep{hwang2025hnet} is
\begin{equation}
\mathcal{L}_{\mathrm{comp}}^{(\ell)}
=\frac{r}{r-1}\left[(r-1)f_{\mathrm h}f_{\mathrm s}
+(1-f_{\mathrm h})(1-f_{\mathrm s})\right].
\label{eq:HiT-budget-original}
\end{equation}
Substituting $\rho=1/r$ and collecting terms gives
Eq.~\eqref{eq:HiT-budget} exactly, with the same loss scale.
We treat $f_{\mathrm h}$ as constant during backpropagation.

Example $b$ enters the stage with $N_b$ tokens and produces $K_b$
groups. Its complete merge tree has $N_b-1$ merges.
Replacing each threshold decision with a sigmoid gives
\begin{equation}
\widetilde{K}_b=N_b-\sum_{m=1}^{N_b-1}
\sigma\!\left(\frac{s_{b,m}-\tau}{T_{\mathrm b}}\right).
\label{eq:HiT-soft-count}
\end{equation}
Here, $s_{b,m}$ is the mean affinity at merge $m$.
The temperature $T_{\mathrm b}$ controls the smoothness of the estimate.
The tree topology is fixed during backpropagation, but its merge
affinities remain differentiable.
The hard retained fraction is $f_{\mathrm h}=\sum_b K_b/\sum_b N_b$.
The soft retained fraction is
$f_{\mathrm s}=\sum_b\widetilde{K}_b/\sum_b N_b$.
At the second stage, $N_b$ is the actual group count from the first
stage. The total compression loss sums the stage losses.

\subsection{Training Details}
\label{sec:training-details}
We use the joint objective in Eq.~\eqref{eq:HiT-objective}.
Given a clean image $x$, we form $x_t=tx+(1-t)\epsilon$,
where $t\in[0,1]$ and $\epsilon$ is Gaussian noise.
Following JiT~\citep{li2025jit}, the network predicts clean images
and uses a velocity loss, equivalently expressed as
\begin{equation}
\mathcal{L}_{\mathrm{diff}}
=
\mathbb{E}_{x,c,t,\epsilon}
\left[
\frac{\operatorname{MSE}
      \bigl(f_\theta(x_t,t,c),x\bigr)}
     {\max(1-t,a_{\min})^2}
\right].
\label{eq:HiT-diffusion-loss}
\end{equation}
Here, $f_\theta$ is the denoiser, $c$ is the conditioning
input, and $a_{\min}=0.05$ is the denominator floor
for both ImageNet and text-to-image generation.

For each forward pass, we sum group counts and token counts across
devices before computing the compression loss.
Gradient accumulation averages these losses over forward passes.

Before training, we calibrate routing thresholds on a fixed set
of noisy training examples to match the target aggregate compression
ratios. Calibration proceeds from outer to inner stages, with each
stage using representations produced by the preceding calibrated
stages. These baseline thresholds remain fixed throughout training
and ordinary sampling. Group guidance uses separately
calibrated inference thresholds
(Appendix~\ref{sec:group-guidance}); for ImageNet, these
follow a precomputed timestep-dependent schedule.

\paragraph{Text-to-image training.}
We pretrain HiT and the MiniT2I baseline on CC12M for
250,000 optimizer updates, then fine-tune on the three-source
mixture from MiniT2I's PyTorch recipe~\citep{wang2026minit2i}:
BLIP3o-60K, DALL-E 3, and ShareGPT-4o-Image.
After filtering, these sources contain 58,859, 19,024, and
41,356 image--caption pairs, respectively (119,239 in total).
We use their released relative sampling weights
$0.06:0.016:0.04$ (equivalently, $15:4:10$).
We retain the released preprocessing and ShareGPT-4o caption
filter, which excludes captions longer than 256 tokens under
the FLAN-T5-Base tokenizer; conditioning still uses
FLAN-T5-Large.
Both models use the same fine-tuning mixture and example order.

The global batch size is 1,024 throughout both phases.
During pretraining, the learning rate increases linearly from
$10^{-6}$ to $4\times10^{-4}$ over the first 5,000 updates
and remains constant thereafter.
Fine-tuning continues at $4\times10^{-4}$ without another
warmup, retaining the optimizer and EMA states.
The SFT40k checkpoint used for HiT in
Table~\ref{tab:text-to-image} is taken after 40,000 fine-tuning
updates, or 290,000 total updates; evaluation uses EMA $0.99995$.

\subsection{Conditioning and Backbone Design}
\label{sec:conditioning}

We start from JiT~\citep{li2025jit}, retaining
RMSNorm~\citep{zhang2019rmsnorm},
query--key normalization~\citep{henry2020qknorm}, and
SwiGLU~\citep{shazeer2020glu}.
Images enter through a single learned linear patch projection,
without the input bottleneck used in JiT.

A key empirical finding is that HiT performs better when class
and time conditioning are supplied entirely through input tokens,
with adaptive layer normalization (adaLN) removed.
We hypothesize that applying separate feature-wise modulation
before, within, and after compression can complicate the interaction
between compact representations and full-resolution skip features.
Token-based conditioning instead carries an evolving conditioning
state through the hierarchy, allowing each stage to incorporate
conditioning through attention.

Following the in-context conditioning formulation explored in
DiT~\citep{peebles2023dit}, we introduce eight class tokens and
one time token at the first Transformer block.
Each class token combines a shared class embedding with a learned
slot embedding, as in JiT~\citep{li2025jit}.
The time token combines a sinusoidal timestep embedding,
transformed by a two-layer SiLU MLP, with a learned token identity
embedding.
These tokens participate in attention throughout the network and
accompany the compact image sequence directly.
They bypass grouping and ungrouping and remain outside the
compression budget and image skip connections.

We retain full-resolution spatial positional encoding, including
image RoPE~\citep{su2024roformer}.
Conditioning tokens receive identity rotations, and compact stages
use no additional RoPE.
The final prediction head operates only on image tokens.
Model widths and stage depths are specified in the experimental
configurations.

\subsection{Joint Grouping for Text-to-Image Generation}
\label{sec:joint-grouping}

Extending the hierarchy to text-to-image generation requires
organizing both image features and variable-length caption content.
We apply grouping jointly to these two modalities.

Following MiniT2I~\citep{wang2026minit2i}, a frozen FLAN-T5-Large
encoder supplies caption features, followed by a learned projection
and two text-only preamble blocks.
The outer encoder and decoder retain separate image and text
projections and MLPs, with joint attention between the modalities.
An explicit time token enters before the first joint block and
accompanies the sequence throughout the hierarchy.

After the outer encoder, we concatenate image features and valid
caption features for grouping.
For example $b$, the eligible population is
\begin{equation}
N_b
=
N_{\mathrm{img}} + \sum_j m_{bj},
\label{eq:HiT-joint-population}
\end{equation}
where $m_{bj}$ is the caption validity mask, including valid special
tokens.
Padding positions are excluded from tree construction, soft pooling,
and compression budgets.
For caption-dropout examples, only image features enter the
grouping path. The time token bypasses grouping and ungrouping at both levels.

The same affinity router and soft aggregation form compact tokens
from the combined population.
A group can contain image features, caption features, or both.
We therefore use shared Transformer blocks within the compact
hierarchy, allowing every group to be processed in the same
representation space.

Ungrouping distributes the processed group features to the original
image and caption positions. We then add the projected skip features.
Padding slots receive their skip features and participate in the
outer decoder as usual.
The restored sequences resume modality-specific processing and
their original positional encodings; only image features are
mapped to the output image.
Both compression budgets use the actual eligible populations,
following Eq.~\eqref{eq:HiT-budget}.

\subsection{Group guidance}\label{sec:group-guidance}

\paragraph{Motivation.}
HiT processes compressed groups through its central backbone,
using fewer tokens than its maximum capacity of one group per
input token. Increasing the number of groups at inference
provides a finer representation, but does not automatically
improve generation: the backbone is trained on compressed
representations, and changing this operating regime can
degrade its predictions. Inspired by Sparse
Guidance~\cite{krause2026sparseguidance}, we use this variation
in token capacity to construct a guidance signal. Specifically,
we combine a conditional prediction with finer grouping and
an unconditional prediction with coarser grouping, using the
same network weights.

\paragraph{Formulation.}
Each grouping stage controls its group population through a
threshold on learned merge scores. We parameterize threshold
adjustments by nominal group-count multipliers
$\mathbf r=(r_1,r_2)$: an adjusted stage targets $r_k$ times
its default mean group count, while $r_k=1$ retains its
original threshold. Group counts remain adaptive for each
input.

Let $v_c^{(\mathbf r)}(x_t,t)$ denote the resulting velocity
prediction under condition $c$. We define group guidance as
\begin{equation}
\begin{aligned}
\hat v_t
&= v_{\varnothing}^{(\mathbf r_u(t))}(x_t,t) \\
&\quad + s\left(
v_c^{(\mathbf r_c(t))}(x_t,t)
- v_{\varnothing}^{(\mathbf r_u(t))}(x_t,t)
\right),
\end{aligned}
\label{eq:group-guidance}
\end{equation}
where $\mathbf r_c(t)\geq\mathbf 1\geq\mathbf r_u(t)$
elementwise. The guidance direction therefore combines
conditioning with a contrast between finer and coarser
representations. Setting both multipliers to $\mathbf 1$
recovers standard classifier-free guidance (CFG).
Group guidance changes only the grouping thresholds,
preserving the learned weights, soft pooling, and readback.
It requires no additional network evaluations beyond CFG.

\paragraph{Calibration.}
We calibrate thresholds for each checkpoint using recorded
merge scores, fitting the modified stages sequentially to
their target mean group counts. For text-to-image generation,
we use 128 fixed image-caption pairs at six noise levels and
fit thresholds shared across timesteps. For ImageNet, we
collect 64 sampling trajectories and fit thresholds separately
in ten equal timestep bins. Calibration requires forward
passes only; all model weights remain fixed.

\paragraph{Text-to-image.}
We use $\mathbf r_c=(1.2,1.2)$ and
$\mathbf r_u=(0.8,1)$ with guidance scale $s=6.5$.
The HiT results in Table~\ref{tab:text-to-image} use the
SFT40k checkpoint trained with the recipe in
Appendix~\ref{sec:training-details}, with EMA $0.99995$ and
100-step Euler sampling. Both CFG and the grouping adjustments
are active throughout $t\in[0,1]$, where $t=0$ denotes pure
noise and $t=1$ a clean image.

\paragraph{ImageNet.}
The 600-epoch HiT result in Table~\ref{tab:imagenet}
(FID 1.48) uses the configuration below: group guidance
with scale $2.6$, EMA $0.9999$, and time-grid shift $k=1.7$.
We use $\mathbf r_c=(1.2,1)$ throughout sampling and apply
additional unconditional compression near the clean end:
\begin{equation}
\mathbf r_u(t)=
\begin{cases}
(1,1),   & t<0.85,\\
(0.5,1), & t\geq0.85.
\end{cases}
\end{equation}
We set $s=2.6$ within the guidance interval $[0.1,1]$,
using the conditional prediction outside this interval.
JiT uses 50 sampling steps on a uniform grid over $t\in[0,1]$,
with Heun updates and a final Euler step.
For this HiT result, we retain the solver and step count
but shift the time grid toward higher noise:
\begin{equation}
\begin{aligned}
\sigma &= 1-t,\qquad k=1.7,\\
\sigma' &= \frac{k\sigma}{1+(k-1)\sigma},
\qquad t'=1-\sigma'.
\end{aligned}
\end{equation}

\paragraph{Convergence comparison.}
The HiT points in Fig.~\ref{fig:hit-teaser} use the group-count
multipliers above with guidance scale $2.6$, while JiT uses CFG $2.4$.
Sampling uses an unshifted time grid except for the 600-epoch
HiT endpoint, which uses $k=1.7$.

\section{Additional Ablations}
\label{sec:additional-ablations}

We further examine compact positional encoding, encoder--decoder
skip fusion, and the compression-loss weight.
All experiments use ImageNet $256\times256$, a patch size
of $16$, and a model width of $768$.
We report ADM FID-50k after 80 training epochs, using
EMA decay $0.9999$, with CFG 2.9 and without group guidance.

The positional-encoding sweep, skip-fusion experiments, and
compression-loss sweep use a 12-block configuration with adaLN,
eight class-context tokens, and linear patch embedding.
Its outer and intermediate encoder/decoder stages each contain
two blocks, while the compact core contains four.
We additionally evaluate the strongest learned positional
variant on our main 18-block HiT-B/16, which uses token
conditioning without adaLN.
Gray table rows identify the corresponding default configurations.

\subsection{Compact Positional Encoding}
\label{sec:ablation-compact-position}

A compact token can represent multiple, potentially disconnected
image regions. We investigate how positional information should
be assigned to these groups.
All variants retain full-resolution image rotary position
embedding (RoPE)~\citep{su2024roformer}.
Table~\ref{tab:compact-position} compares fixed RoPE rules,
learned RoPE, and additional size or embedding components.

\paragraph{Fixed RoPE.}
We first compare two ways to construct group-level RoPE:
evaluating it at a representative coordinate, or averaging
the constituent patches' rotary tables.
For original patch coordinates $p_i$, the hard centroid
uses uniform weights over hard members, while the soft
centroid uses normalized pooling weights:
\begin{equation}
    \mu_g^{\mathrm{hard}}
    = \frac{1}{|G_g|}\sum_{i\in G_g}p_i,
    \qquad
    \mu_g^{\mathrm{soft}}
    = \sum_i P_{ig}p_i.
\end{equation}
The \emph{hard centroid} and \emph{soft centroid} variants
evaluate ordinary two-dimensional RoPE at these coordinates.
The \emph{hard table average} and \emph{soft table average}
instead average the original cosine and sine tables:
\begin{equation}
    C_g = \sum_i w_{ig}\cos\theta(p_i),
    \qquad
    S_g = \sum_i w_{ig}\sin\theta(p_i),
\end{equation}
where $\theta(p_i)$ is the original rotary phase and
$w_{ig}$ uses either uniform hard-member weights or
normalized soft pooling weights.
At the second grouping level, hard supports are unions
of constituent groups and soft weights are composed
across pooling levels, preserving the original patch
coordinate system.
Among these four choices, the hard centroid gives the
lowest FID, improving the 12-block control from $12.27$
to $11.35$.

\paragraph{Learned RoPE.}
We next replace the fixed positional rule with learned
parameters. A stage-specific coefficient mixes hard and
soft centroids and second moments.
The mixed centroid determines the rotary phase, while
the resulting spatial variance attenuates each frequency
by $\exp(-\gamma\omega^2v_g/2)$, where $\omega$ is the
frequency, $v_g$ is the positional variance, and
$\gamma$ is a learned positive scale.
A learned coefficient also controls a logarithmic
group-size bias in attention.
These parameters are shared across groups, with
head- and frequency-specific settings where applicable.

The \emph{per-group MLP} is an additional module that
predicts adjustments to these parameters from group
statistics and class/time conditioning.
It has two hidden layers of width 64 and a zero-initialized
output layer, so it initially reproduces the learned
RoPE rule without the MLP.
The complete learned RoPE plus MLP adds 66,700 parameters
across both grouping levels.
Positional statistics are detached from the router,
while the added parameters are optimized through the
generation objective.
Adding the MLP changes FID from $11.48$ to $11.18$
on the 12-block model.
The same configuration improves the main 18-block
model from $6.61$ to $6.25$.

\paragraph{Size and embedding extensions.}
We consider three additional components.
\emph{Size frequencies} replace the four slowest rotary
frequencies along each spatial axis with an encoding
of $\log_2|G_g|$.
\emph{Log-size bias} adds $\log|G_g|$ to the attention
logits for group $g$ as a key, following the proportional
attention formulation of ToMe~\citep{bolya2023tome}.
A \emph{learned positional embedding (LPE)} provides
an additional learned group-position representation
for compact attention; we test it both alone and
alongside hard-centroid RoPE.
The log-size bias and LPE slightly improve the
hard-centroid result, whereas size frequencies weaken it.
LPE alone provides little improvement over the control.
Overall, performance remains relatively close across
the tested positional variants.
For simplicity, we omit compact-stage RoPE from our
default configuration.

\begin{table}[t]
    \centering
    \small
    \setlength{\tabcolsep}{3pt}
    \renewcommand{\arraystretch}{1.10}
    \begin{tabular*}{\columnwidth}{
        @{\extracolsep{\fill}}llr
    }
        \toprule
        \textbf{Compact RoPE}
        & \textbf{Addition}
        & \textbf{FID$\downarrow$} \\
        \midrule
        \multicolumn{3}{@{}l}{
            \textit{HiT-B/16 (adaLN, D12)}
        } \\

        Hard centroid
        & -- & 11.35 \\
        Soft centroid
        & -- & 11.48 \\
        Hard table average
        & -- & 11.74 \\
        Soft table average
        & -- & 11.96 \\

        \addlinespace[4pt]
        Learned RoPE
        & -- & 11.48 \\
        Learned RoPE
        & Per-group MLP & \textbf{11.18} \\

        \addlinespace[4pt]
        Hard centroid
        & Size frequencies & 12.03 \\
        Hard centroid
        & Log-size bias & 11.23 \\
        Hard centroid
        & LPE & 11.24 \\
        None
        & LPE & 12.13 \\

        \midrule
        \HiTrowshade
        None (default)
        & -- & 12.27 \\

        \midrule
        \multicolumn{3}{@{}l}{
            \textit{HiT-B/16 (In-context, D18)}
        } \\
        Learned RoPE
        & Per-group MLP & \textbf{6.25} \\
        \midrule
       \HiTrowshade
        None (default)
        & -- & 6.61 \\
        \bottomrule
    \end{tabular*}
    \caption{
        \textbf{Compact positional encoding.}
        We compare positional encodings for grouped tokens
        at the compact stages.
        D12/D18 denote 12/18 Transformer blocks;
        LPE denotes learned positional embedding.
        Highlighted rows are the corresponding defaults.
    }
    \label{tab:compact-position}
\end{table}

\subsection{Skip Fusion}
\label{sec:ablation-skip-fusion}

Skip connections combine fine-resolution encoder features
with information processed through the compact hierarchy.
We evaluate the skip-fusion rules studied in
U-ViT~\citep{bao2023uvit} within HiT's hierarchy,
examining whether their parameterization affects
generation quality.
This ablation uses HiT-B/16 with adaLN and
12 Transformer blocks.

Let $H$ denote the encoder features before grouping and
$R$ the reconstructed features returned by ungrouping at
the same resolution.
Our default uses a learned projection on the skip branch:
\begin{equation}
    H_{\mathrm{out}} = R + HW_{\mathrm{s}}^\top.
\end{equation}
We compare this with plain addition, concatenation
followed by projection, and addition followed by projection:
\begin{align}
    H_{\mathrm{out}} &= R + H, \\
    H_{\mathrm{out}} &= [R\,\|\,H]W_{\mathrm{c}}^\top, \\
    H_{\mathrm{out}} &= (R+H)W_{\mathrm{f}}^\top,
\end{align}
where $\|$ denotes concatenation along the feature dimension.
The square projections are initialized to the identity,
and the concatenation projection is initialized as
$W_{\mathrm{c}}=[I\ I]$.
Thus, all four parameterizations initially compute $R+H$.
Each choice is applied at both hierarchy levels, with
grouping, soft pooling, and ungrouping unchanged.

Table~\ref{tab:skip-fusion} shows that plain addition
achieves the lowest FID, followed by concatenation
and projection.
Both remain relatively close to the default projected
skip, whereas projecting the sum degrades performance.
More elaborate fusion therefore does not consistently
improve generation quality.
We retain the projected skip as a simple fusion rule
that allows the encoder contribution to be transformed
while preserving a direct path for reconstructed
group features.

\begin{table}[t]
    \centering
    \small
    \setlength{\tabcolsep}{3pt}
    \renewcommand{\arraystretch}{1.10}
    \begin{tabular*}{\columnwidth}{
        @{\extracolsep{\fill}}llr@{}
    }
        \toprule
        \textbf{Fusion}
        & \textbf{Operation}
        & \textbf{FID$\downarrow$} \\
        \midrule
        \multicolumn{3}{@{}l}{
            \textit{HiT-B/16 (adaLN, D12)}
        } \\
        Plain addition
        & $R+H$
        & \textbf{11.65} \\
        Concat.\ + proj.
        & $[R\,\|\,H]W_{\mathrm{c}}^\top$
        & 12.00 \\
        Add, then project
        & $(R+H)W_{\mathrm{f}}^\top$
        & 13.34 \\
        \midrule
        \HiTrowshade
        Projected skip
        & $R+HW_{\mathrm{s}}^\top$
        & 12.27 \\
        \bottomrule
    \end{tabular*}
    \caption{
        \textbf{Skip fusion.}
        We compare ways to fuse ungrouped features with
        the stage skip.
        D12 denotes 12 Transformer blocks.
        Fusion changes apply at both hierarchy levels;
        the highlighted row is the default.
    }
    \label{tab:skip-fusion}
\end{table}

\subsection{Compression-Loss Weight}
\label{sec:ablation-budget-weight}

The compression loss regulates the number of compact
tokens used by the hierarchy.
We vary its weight to examine whether generation quality
requires precise tuning of this regularization.
This ablation uses HiT-B/16 with adaLN and
12 Transformer blocks.

For the two-stage model, the objective is
\begin{equation}
    \mathcal{L}
    = \mathcal{L}_{\mathrm{diffusion}}
    + \lambda
      \left(
        \mathcal{L}_{\mathrm{comp}}^{(1)}
        + \mathcal{L}_{\mathrm{comp}}^{(2)}
      \right).
\end{equation}
The coefficient $\lambda$ multiplies the sum of stage losses.
We evaluate $\lambda\in\{0.005,0.01,0.02,0.03\}$,
retaining the $2\times$ compression target at each stage
and the remaining model and training settings.

Table~\ref{tab:budget-weight} shows that FID ranges
from $12.07$ to $12.33$ across this sixfold range of weights.
The setting $\lambda=0.02$ gives the lowest observed FID,
with a small improvement over $\lambda=0.01$.
Generation quality is therefore relatively insensitive
to this hyperparameter over the tested range,
and we retain $\lambda=0.01$ as the default.

\begin{table}[t]
    \centering
    \small
    \setlength{\tabcolsep}{25pt}
    \renewcommand{\arraystretch}{1.10}
    \begin{tabular}{@{}lr@{}}
        \toprule
        \textbf{Weight $\lambda$}
        & \textbf{FID$\downarrow$} \\
        \midrule
        \multicolumn{2}{@{}l@{}}{
            \textit{HiT-B/16 (adaLN, D12)}
        } \\
        $0.005$ & 12.30 \\
        $0.02$ & \textbf{12.07} \\
        $0.03$ & 12.33 \\
\midrule
\multicolumn{1}{
    @{}>{\columncolor{gray!12}[0pt][\tabcolsep]}l
}{$0.01$ (default)}
&
\multicolumn{1}{
    >{\columncolor{gray!12}[\tabcolsep][0pt]}r@{}
}{12.27} \\
\bottomrule
    \end{tabular}
    \caption{
        \textbf{Compression-loss weight.}
        We vary the weight $\lambda$ of the compression loss.
        The highlighted row is the default.
    }
    \label{tab:budget-weight}
\end{table}

\subsection{Number of Grouping Stages}
\label{sec:ablation-hierarchy-depth}

We examine whether adding a third grouping stage improves
generation quality.
All variants use HiT-B/16 with in-context conditioning,
width $768$, and 18 Transformer blocks.
Starting from the default two-stage layout
$3/3/6/3/3$, we reorganize its six central blocks into
a third encoder, bottleneck, and decoder with two
blocks each, giving $3/3/2/2/2/3/3$.
The additional level introduces another pair of grouping and ungrouping operations
and a projected skip connection, increasing the parameter
count from $133.9$M to $135.6$M.

We evaluate compression ratios of $2\times2\times2$
and $2\times2\times4$, targeting average image-token
populations of $256\rightarrow128\rightarrow64\rightarrow32$
and $256\rightarrow128\rightarrow64\rightarrow16$,
respectively.
Group counts remain input-dependent.
Each stage retains a compression-loss weight of $0.01$.
We also test the layout $2/2/2/6/2/2/2$, which allocates
more blocks to the deepest bottleneck.
Layout entries specify block counts in execution order,
from the outer encoder through the bottleneck to the
outer decoder.

As shown in Table~\ref{tab:hierarchy-depth}, the
$3/3/2/2/2/3/3$ configuration achieves FID $6.66$
with $2\times2\times2$ compression, close to the
two-stage default's $6.61$.
Stronger compression at the third stage gives $6.95$,
while concentrating more blocks in the deepest stage
gives $8.82$.
Since the best three-stage result is close to the
two-stage result, we retain two grouping stages
by default for architectural simplicity.

\begin{table}[t]
    \centering
    \small
    \setlength{\tabcolsep}{3pt}
    \renewcommand{\arraystretch}{1.10}
    \begin{tabular*}{\columnwidth}{
        @{\extracolsep{\fill}}cllrr@{}
    }
        \toprule
        \textbf{Stages}
        & \textbf{Compression}
        & \textbf{Block layout}
        & \textbf{GFLOPs}
        & \textbf{FID$\downarrow$} \\
        \midrule
        \multicolumn{5}{@{}l@{}}{
            \textit{HiT-B/16 (In-context, D18)}
        } \\
        3 & $2\times2\times2$
        & $3/3/2/2/2/3/3$ & 21.9 & 6.66 \\
        3 & $2\times2\times4$
        & $3/3/2/2/2/3/3$ & 21.6 & 6.95 \\
        3 & $2\times2\times2$
        & $2/2/2/6/2/2/2$ & 17.1 & 8.82 \\
        \midrule
        \HiTrowshade
        \textbf{2}
        & $\boldsymbol{2\times2}$
        & $\boldsymbol{3/3/6/3/3}$
        & \textbf{22.2}
        & \textbf{6.61} \\
        \bottomrule
    \end{tabular*}
    \caption{
        \textbf{Number of grouping stages.}
        We compare two and three grouping stages
        at 18 blocks.
        The highlighted row is the default.
    }
    \label{tab:hierarchy-depth}
\end{table}

\section{Generated Samples}
\label{sec:generated-samples}
\begin{figure*}[p]
    \centering
    \includegraphics[width=\textwidth]{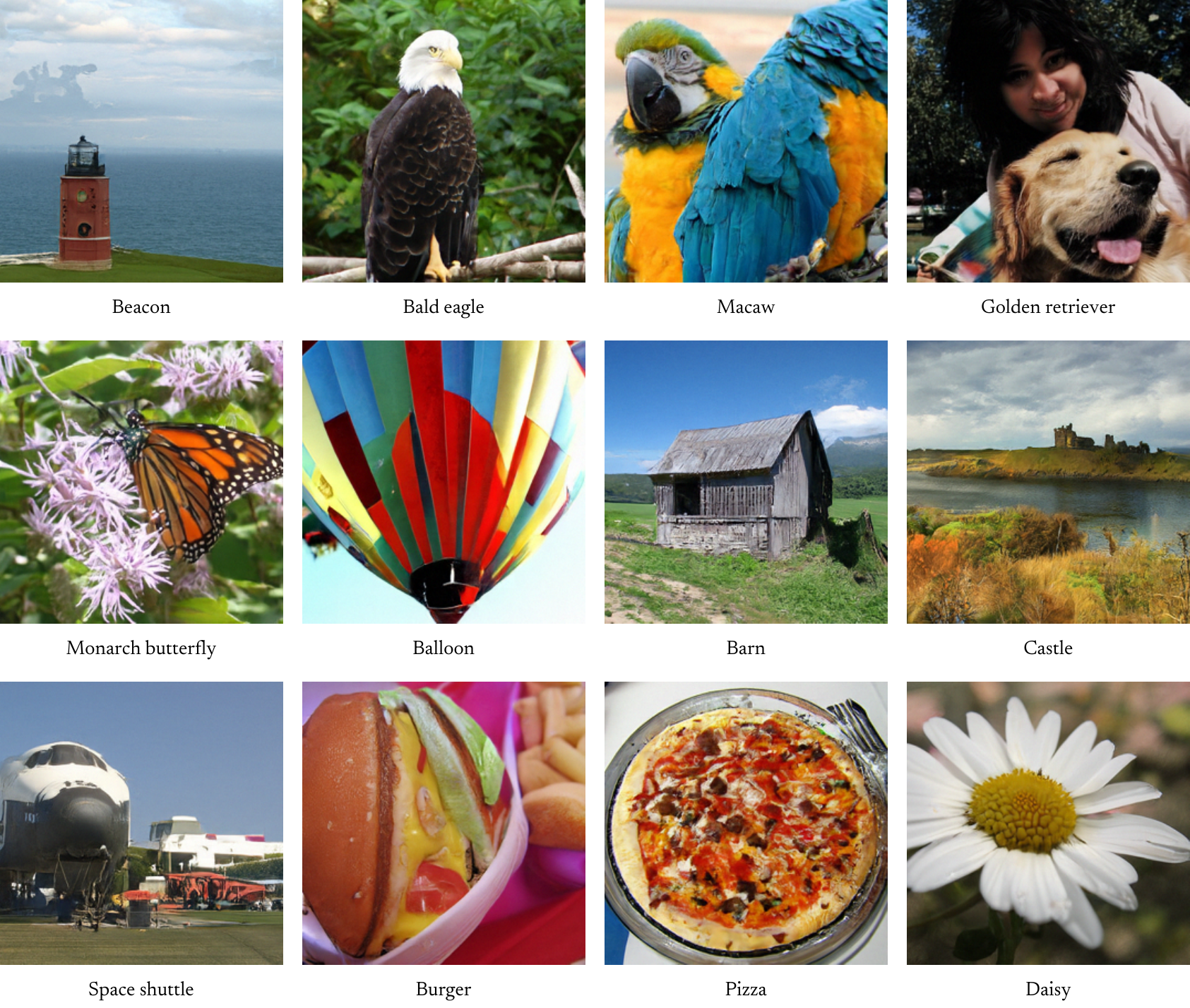}
    \caption{\textbf{ImageNet generation.}
    Twelve selected $256\times256$ samples from HiT-XL/16 at 600 epochs,
    using EMA $0.9999$, guidance scale $2.6$, and group guidance.
    Sampling uses 50 Heun steps with a final Euler step and time shift $1.7$.
    Each label gives the conditioning class.}
    \label{fig:hit-generated-imagenet}
\end{figure*}

\begin{figure*}[p]
    \centering
    \includegraphics[width=\textwidth]{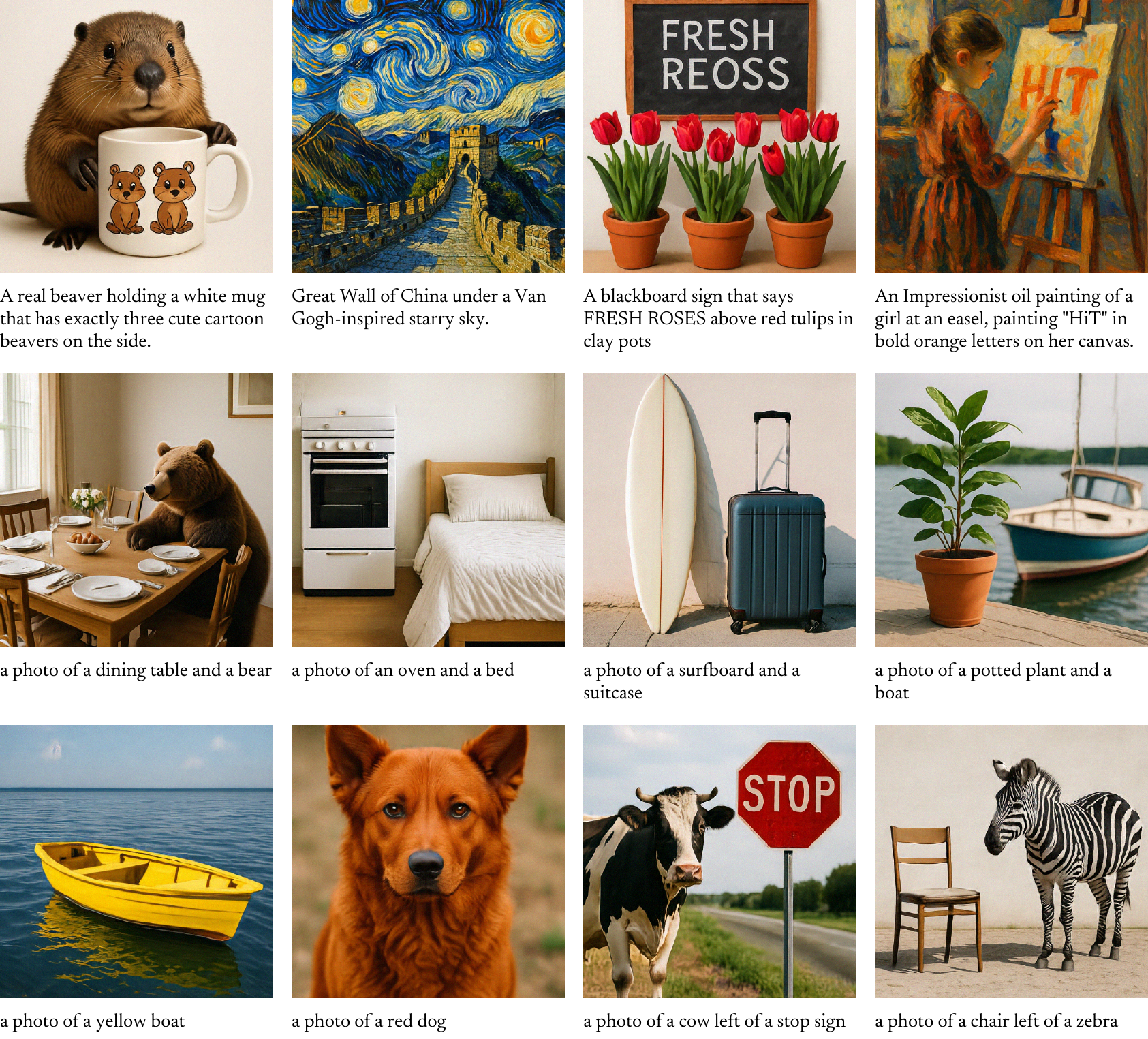}
    \caption{\textbf{Text conditioned image generation.}
    Twelve selected $512\times512$ samples from HiT-B/16 D27,
    using EMA $0.99995$ after 250k pretraining and 40k fine tuning updates.
    Sampling uses 100 Euler steps with CFG $7.0$.
    Each image is paired with its full prompt.}
    \label{fig:hit-generated-t2i}
\end{figure*}

Figure~\ref{fig:hit-generated-imagenet} shows ImageNet samples from
HiT-XL/16 at 600 epochs.
Figure~\ref{fig:hit-generated-t2i} shows text conditioned samples from
HiT-B/16 D27 after 250k pretraining and 40k fine tuning updates.

\section{Additional Visualizations}
\label{sec:additional-visualizations}
\paragraph{Visualization setup.}
The following routing figures inspect HiT-B/16 D27 after 250k
pretraining and 40k fine tuning updates, using EMA $0.99995$ and
held out COCO image and caption pairs. Its $512\times512$ inputs
use $16\times16$ pixel patches on a $32\times32$ grid.
Image patches and valid caption tokens share groups.
Padding and the time token are excluded.
The spatial maps show the image portion of these joint groups.
Unless stated otherwise, they show stage 2 groups on clean inputs
at $t=1$.

\subsection{Image Groups}
\begin{figure*}[p]
    \centering
    \includegraphics[width=\textwidth]{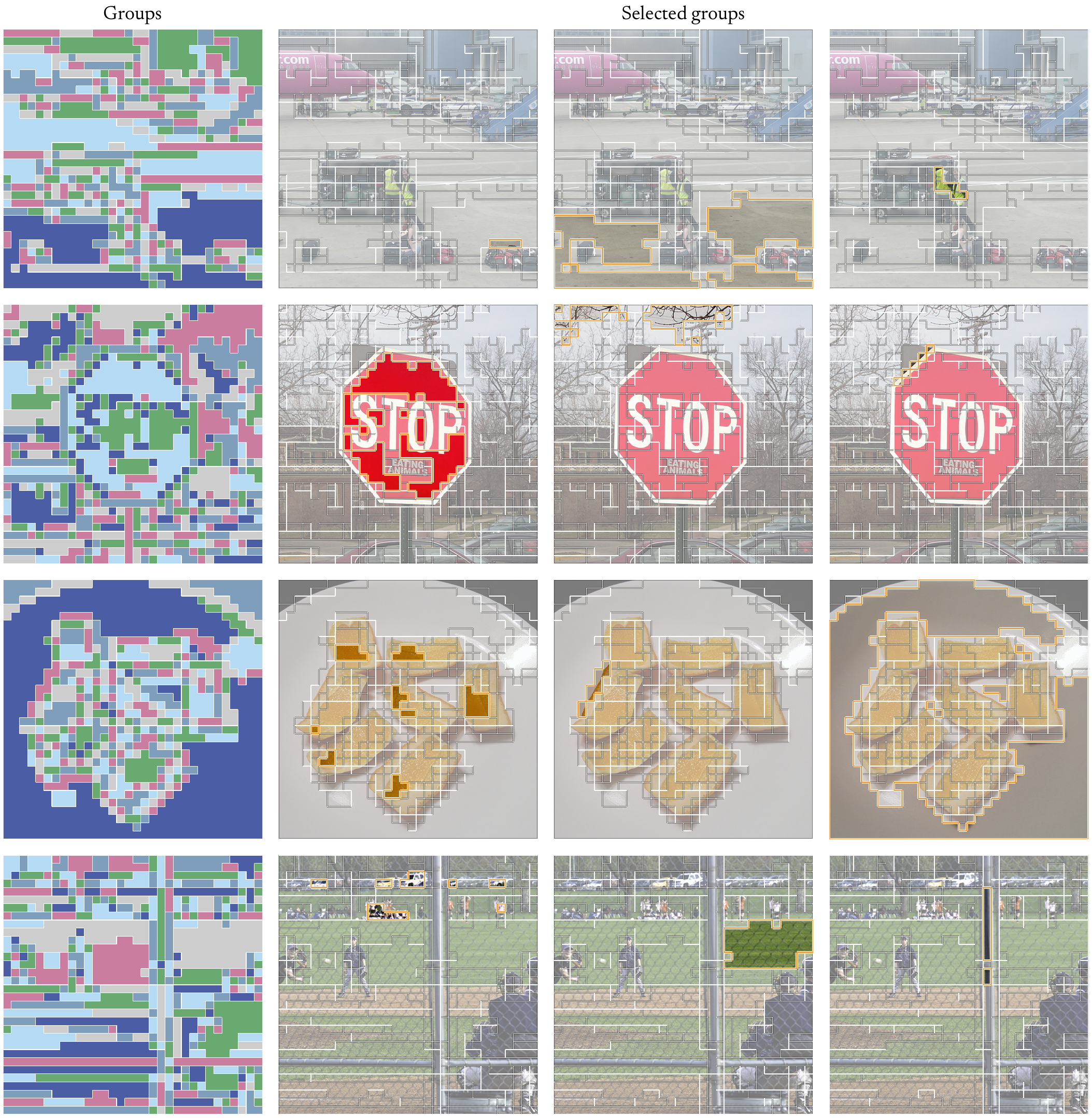}
    \caption{\textbf{Image groups at stage 2.}
    All examples use clean inputs at $t=1$.
    Rows show an airport scene, a stop sign, orange peels, and a baseball scene.
    The left column shows the full group map.
    Each of the three remaining panels highlights one selected group in yellow.
    White lines show all hard group boundaries.
    Colors distinguish touching groups and can repeat in distant groups.}
    \label{fig:t2i-image-groups}
\end{figure*}

Figure~\ref{fig:t2i-image-groups} shows groups that cover object regions,
background regions, and spatially separated patches.

\subsection{Grouping Across HiT Stages}
\begin{figure*}[p]
    \centering
    \includegraphics[width=\textwidth]{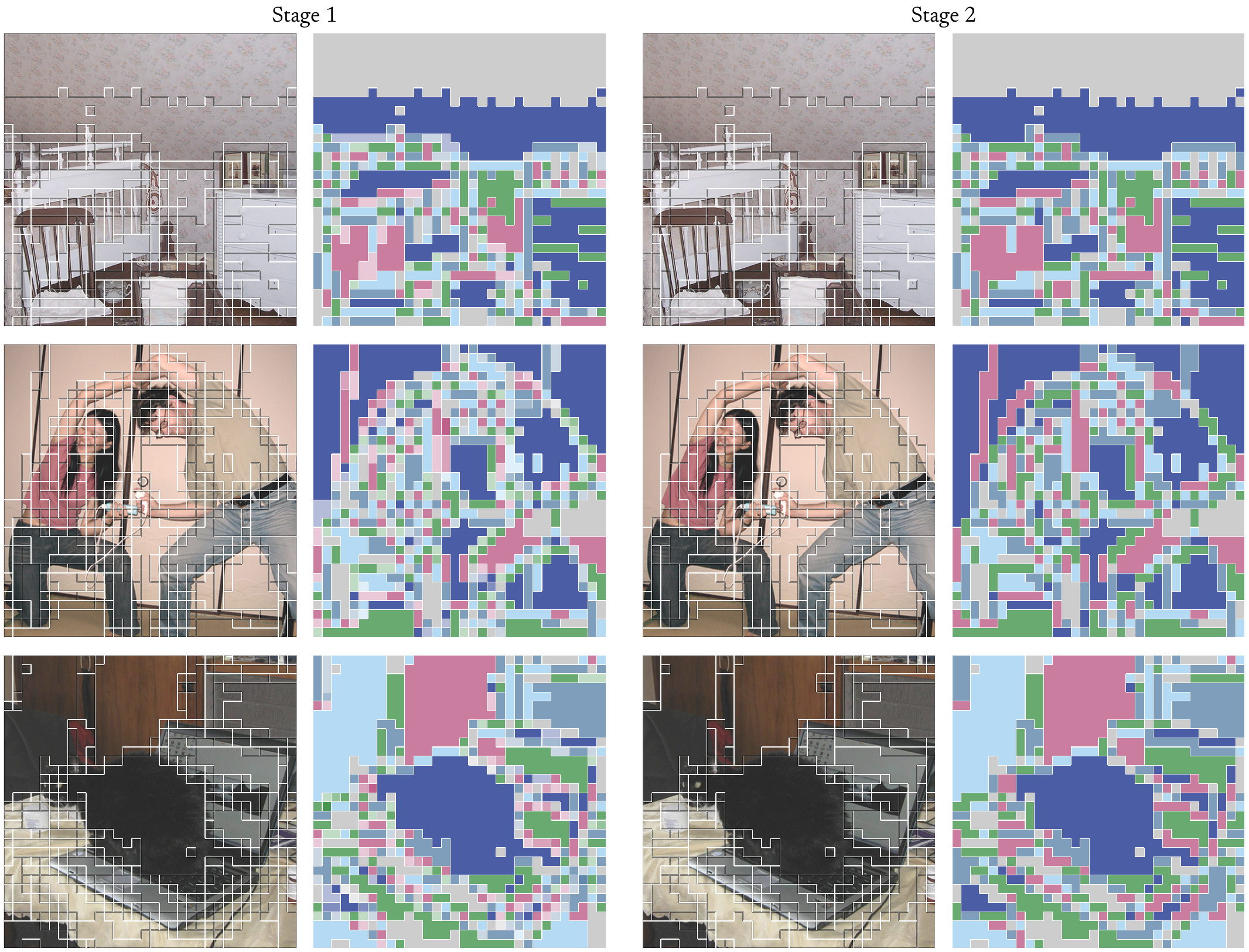}
    \caption{\textbf{Grouping across HiT stages.}
    Rows show a bedroom, people playing a game, and a cat on a laptop.
    Each stage shows the clean image with group boundaries and its group map
    at $t=1$.
    Related shades mark first stage groups with the same second stage parent.
    White boundaries distinguish the groups at each stage.
    Colors can repeat for distant parents.}
    \label{fig:hit-stage-comparison}
    \label{fig:t2i-hierarchy}
\end{figure*}

The second grouping stage operates on the compact features produced
by the first stage. Under the hard assignments, each first-stage
group has one second-stage parent.
Figure~\ref{fig:hit-stage-comparison} visualizes this hierarchy
on the image grid. Pooling and ungrouping remain soft, allowing
feature exchange beyond the displayed hard memberships.

\subsection{Grouping Across Diffusion Times}
\begin{figure*}[p]
    \centering
    \includegraphics[width=\textwidth]{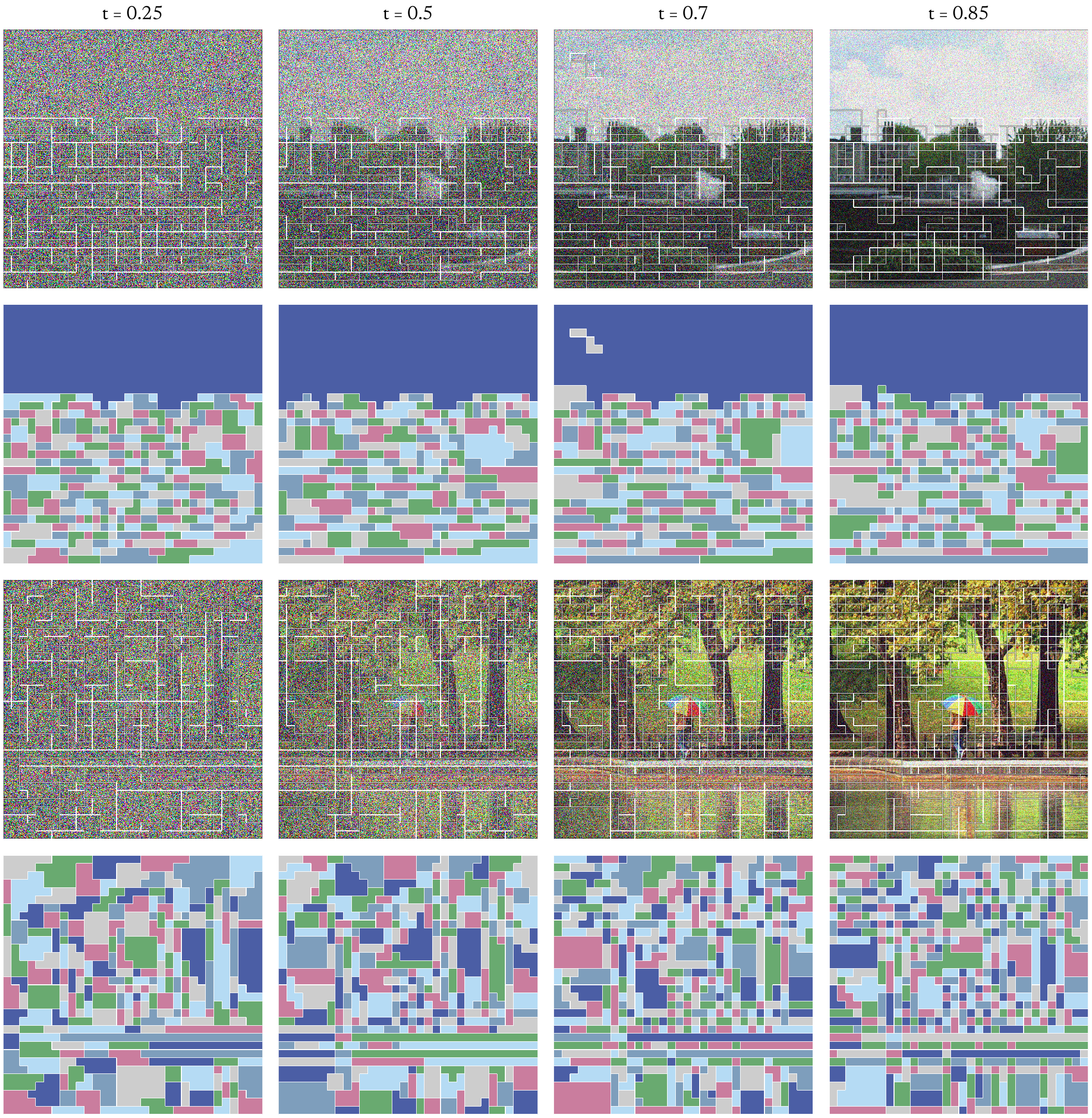}
    \caption{\textbf{Grouping across diffusion times.}
    Columns use $t=0.25$, $0.5$, $0.7$, and $0.85$ at stage 2.
    Each pair of rows shows the noisy image with group boundaries above its group map.
    The train and umbrella examples use one fixed noise draw per image.
    Colors distinguish groups within each map and do not track group identity
    across times.}
    \label{fig:hit-time-comparison}
\end{figure*}

Figure~\ref{fig:hit-time-comparison} holds the image, caption, and noise
realization fixed while changing $t$ in $x_t=tx+(1-t)\epsilon$.
The model computes groups from each noisy input.
The displayed inputs use the noisy image views saved with the website export.

\subsection{Learned Image--Text Associations}
\label{app:cross-modal}
\begin{figure*}[p]
    \centering
    \includegraphics[width=\textwidth]{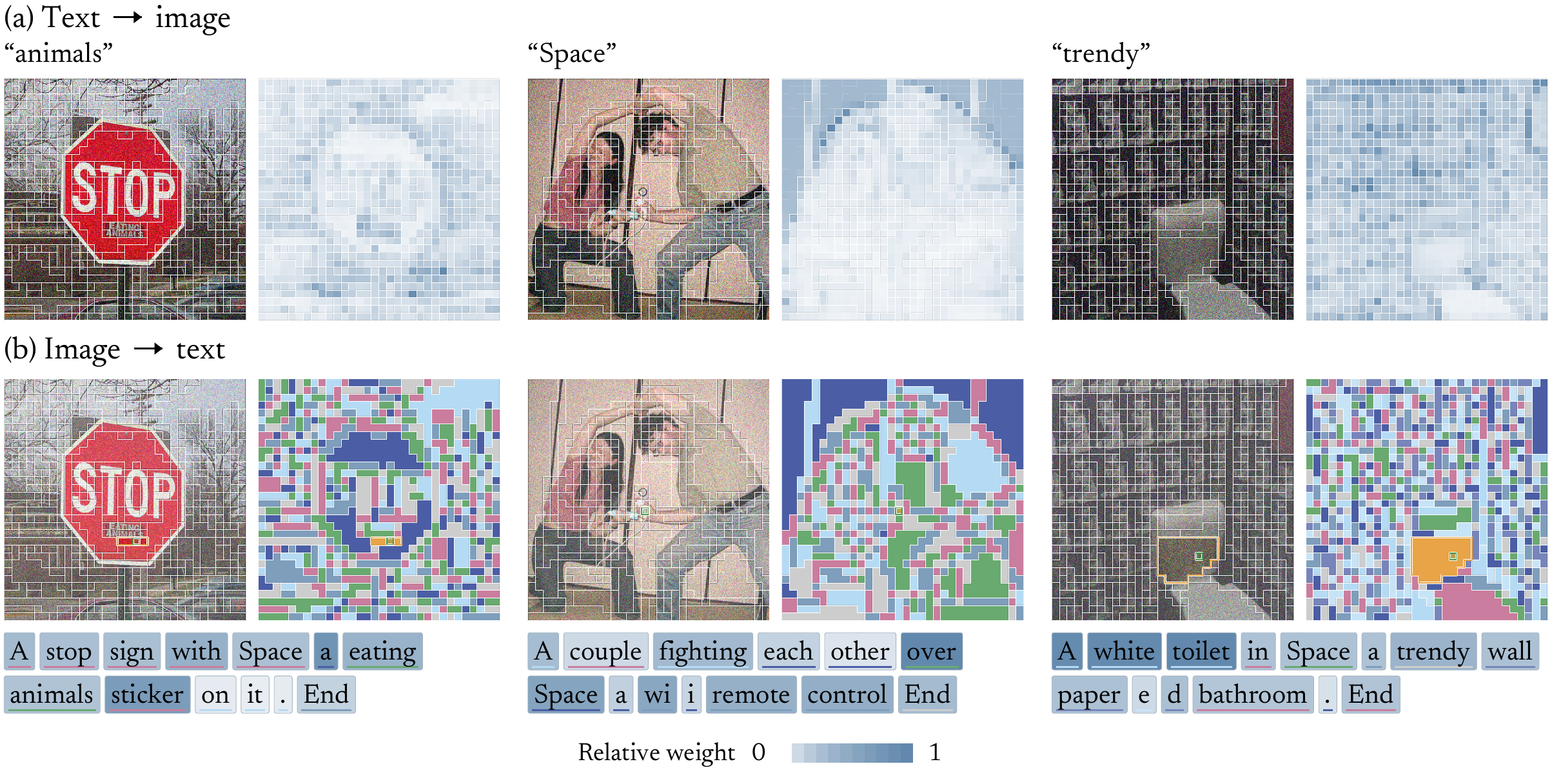}
    \caption{\textbf{Text and image connections at stage 2.}
    All examples use $t=0.85$.
    (a) The top row shows image pooling weights for the groups that contain
    ``animals'', the word start token ``Space'', and ``trendy''.
    (b) The bottom row shows the corresponding image to text views.
    Green marks the selected image patch. Yellow marks its hard group.
    Blue caption tokens show soft routing weights from the selected patch.
    White lines show all hard group boundaries.
    Group colors distinguish touching groups and can repeat in distant groups.
    Each query or selected patch uses a shared maximum across both stages
    to normalize its weights.
    Both displays use a square root color scale.}
    \label{fig:hit-text-connections}
    \label{fig:additional-cross-modal}
    \label{fig:teaser-routing}
    \label{fig:t2i-pooling-both}
    \label{fig:t2i-reverse-routing}
\end{figure*}

Figure~\ref{fig:hit-text-connections} shows connections between caption
features and image features at stage 2.
A caption query selects an image pooling map.
A selected image patch gives soft connections to caption tokens.

\subsection{Routing Visualization Details}
\label{sec:routing-visualization-details}
Let $A_\ell$ contain the row normalized ungrouping weights and let
$(P_\ell)_{ig}=(A_\ell)_{ig}/\sum_j(A_\ell)_{jg}$ contain the pooling weights.
At the first stage, token indices span the joint population of image
patches and valid caption tokens; at the second, they index the
first-stage groups. Padding and the time token are excluded.
For a caption token with stage 2 hard group $g$, the query view uses
$(P_1P_2)_{\mathrm{img},g}$.
The website export stores the routing weights in FP32.
For each caption query, we normalize image weights by their maximum
across both stages and apply a square root color scale.
We retain the patch grid without spatial smoothing.
Caption tokens in the same hard group use the same image pooling map.

For image patch $i$, the caption view uses the caption entries of
$((A_1A_2)(P_1P_2)^\top)[i,:]$.
Each token piece retains its saved weight. The display includes separate
word start tokens, punctuation, and the caption end token.
We divide these token weights by their maximum across both stages for the
same selected patch, then apply the square root color scale.
These weights describe the pooling and ungrouping paths through the model.

\paragraph{Main figure.}
Figure~\ref{fig:hit-teaser} uses the same checkpoint and FP32 export.
It shows stage 2 groups on a clean input at $t=1$.
The selected patch is at row 13, column 26, counting from zero.
The selected caption word is ``boy''.

\end{document}